\documentclass[10pt,twocolumn,letterpaper]{article}

\usepackage{wacv}
\usepackage{times}
\usepackage{epsfig}
\usepackage{graphicx}
\usepackage{amsmath}
\usepackage{amssymb}
\usepackage{mathtools}

\usepackage{subfig}
\newcommand{\centered}[1]{\begin{tabular}{l} #1 \end{tabular}}
\usepackage{comment}

\newcommand{\boldparagraph}[1]{\vspace{0.2cm}\noindent{\bf #1:} }

\def\wacvPaperID{1072} 

\wacvfinalcopy 
\ifwacvfinal
\def\assignedStartPage{1} 

\fi

\ifwacvfinal
\usepackage[breaklinks=true,colorlinks,bookmarks=false]{hyperref}
\else
\usepackage[pagebackref=true,breaklinks=true,colorlinks,bookmarks=false]{hyperref}
\fi

\ifwacvfinal
\else
\fi

\begin{document}

\title{Dynamic Plane Convolutional Occupancy Networks}

\author{Stefan Lionar$^{1}$\thanks{Equal contribution. This is a 3D Vision course project at ETH Zurich.}
\qquad Daniil Emtsev$^{1}$\footnotemark[1]
\qquad Dusan Svilarkovic$^{1}$\footnotemark[1]
\qquad Songyou Peng$^{1, 2}$\\
$^1$ETH Zurich \quad $^2$ Max Planck ETH Center for Learning Systems\\
{\tt\small \{slionar, demtsev, dsvilarko\}@ethz.ch \quad songyou.peng@inf.ethz.ch}
}

\maketitle

\begin{abstract}

Learning-based 3D reconstruction using implicit neural representations has shown promising progress not only at the object level but also in more complicated scenes. In this paper, we propose Dynamic Plane Convolutional Occupancy Networks, a novel implicit representation pushing further the quality of 3D surface reconstruction. The input noisy point clouds are encoded into per-point features that are projected onto multiple 2D dynamic planes. A fully-connected network learns to predict plane parameters that best describe the shapes of objects or scenes.
To further exploit translational equivariance, convolutional neural networks are applied to process the plane features. Our method shows superior performance in surface reconstruction from unoriented point clouds in ShapeNet as well as an indoor scene dataset. Moreover, we also provide interesting observations on the distribution of learned dynamic planes.

\end{abstract}

\begin{figure*}[ht!]
\centering
\includegraphics[width=1.0\textwidth]{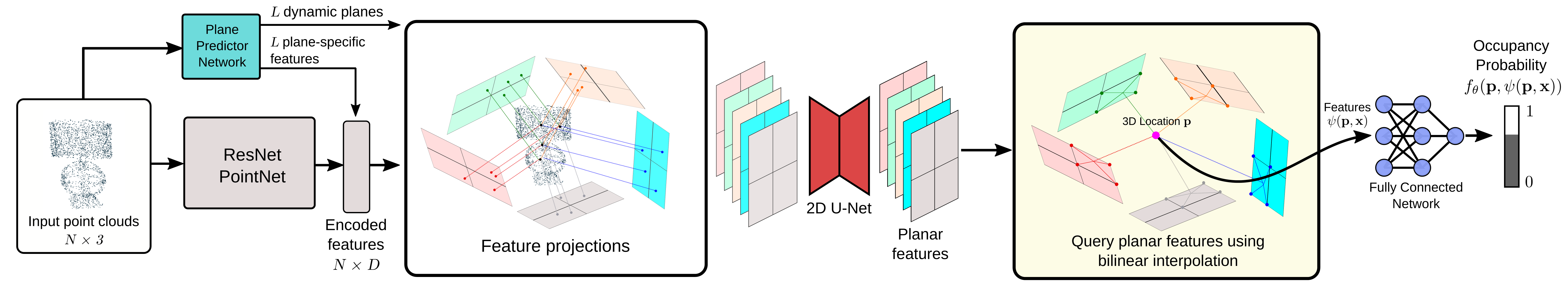}
\caption{\textbf{Dynamic Plane Convolutional Occupancy Networks pipeline.} $N$ input point clouds are encoded to per-point features by ResNet PointNet~\cite{pointnet} with $D$ as the feature dimension. Concurrently, a shallow plane predictor network learns $L$ dynamic planes and plane-specific features from the input point clouds. We sum the plane-specific features to all of the encoded per-point features with respect to individual dynamic planes. Next, the summed features are projected to the dynamic planes. The projected plane features are then processed using U-Net~\cite{unet} with shared weights among planes. In the decoding phase, the occupancy of a uniformly sampled point $\mathbf{p}$ is predicted by a shallow fully-connected network conditioned on the queried local planar features.}
\label{fig:method}
\end{figure*}

\section{Introduction}\label{sec:intro}

Exploiting 3D information, such as point clouds, has become increasingly popular for various computer vision tasks, such as self-driving vehicles, indoor navigation, and robotics~\cite{boss, indoor}. 3D surface reconstruction from point clouds promises better precision for these applications. In recent years, learning-based 3D reconstruction using implicit neural representations in the continuous domain has gained much attention due to its ability to produce smooth and expressive reconstruction with a significant reduction of memory footprint \cite{chibane2020implicit, mescheder2018occupancy, peng2020convolutional}.

However, the pioneering implicit 3D reconstruction approaches are limited to single objects and do not scale to larger scenes due to the use of global embeddings. Some recent works~\cite{chabra2020deep, chibane2020implicit, genova2020local, jiang2020local, peng2020convolutional} noticed this problem and introduced various ways to exploit the information of local structures. While those works have introduced a significant improvement in the object or scene-level reconstruction, the work of Peng \etal on convolutional occupancy networks (ConvONet)~\cite{peng2020convolutional} is the first to demonstrate an accurate and efficient reconstruction of large-scale scenes from point clouds without the need for online optimization. 
One critical success factor of this work is encoding 3D inputs to 2D canonical planes, which are then processed by convolutional neural networks (CNNs).
In this way, the translation equivariant property of CNNs and the local similarity of 3D structures are exploited, enabling the accurate reconstruction of complex scenes. In their work, three canonical planes are pre-defined following Manhattan-world assumption~\cite{coughlan1999manhattan} on the dataset orientation.

In this work, we propose \emph{Dynamic Plane Convolutional Occupancy Networks}\footnote{
Code is available at: \url{https://github.com/dsvilarkovic/dynamic_plane_convolutional_onet}.}, an implicit representation that enables accurate scene-level reconstruction from 3D point clouds. Instead of learning features on three pre-defined canonical planes as in~\cite{peng2020convolutional}, we use a fully-connected network to learn dynamic planes, on which we project the encoded per-point features. The learned dynamic planes capture rich features over the most informative directions. We systematically investigate the use of up to 7 learned planes and demonstrate progressive improvements by increasing the number of learned planes in our experiment. The detailed architecture of our model is illustrated in Fig.~\ref{fig:method}. Compared to ~\cite{peng2020convolutional}, our model introduces another degree of precision by learning features that are more specific to every object and plane.

In summary, the main contributions of our paper are as follows:
\begin{itemize}
    \item We fully leverage deep neural networks in feature learning to predict the best planes for 3D surface reconstruction tasks from unoriented point clouds. 
    \item We show superiority over state-of-the-art approaches in the task of 3D surface reconstruction at both object and scene level.
    \item We provide various observations on the distribution of the dynamic planes from intensive experiments.
\end{itemize}

In addition, we exploit the use of positional encoding proposed in~\cite{mildenhall2020nerf}, which maps the low dimensional 3D point coordinates to higher-dimension representations with periodic functions under various frequencies. While \cite{mildenhall2020nerf} shows its effectiveness on image rendering tasks, we demonstrate that the same positional encoding is also useful for 3D reconstruction tasks. 

Additionally, we formulate a similarity loss function to govern the orientations of dynamic planes to orient in diverse directions. By using a high number of dynamic planes trained with the additional similarity loss function, we observe a considerable improvement in the reconstruction from point clouds with unseen orientations.

\section{Related Work}\label{sec:related_work}
\par
Existing works on learning-based 3D reconstruction can be broadly categorized by the output representation: voxels, points, meshes, or implicit representations.

\textbf{Voxel representations:} 
Voxel might be the most widely used representation for 3D reconstruction~\cite{choy20163d, wu2016learning, wu20153d}, but is limited in terms of resolution due to its large memory consumption. To alleviate this problem, several works consider the multi-scale scheme or octrees for efficient space partitioning~\cite{dai2017shape,hane2017hierarchical, maturana2015voxnet}. Even with these modifications, these approaches are still restricted by computation and memory.

\textbf{Point representations:}
Point clouds are widely used either in robotics or computer graphics~\cite{fan2017point,pointnet,pointnet++}. 
However, there are no topological relations among points, so extra post-processing steps are required~\cite{curless1996volumetric,dai2020sg}. Several works also propose learning-based convolution operations on point clouds to describe the relation among points, analogous to the 2D convolution on image pixels in convolutional neural networks~\cite{lin2020fpconv, point-planenet, thomas2019kpconv, wu2019pointconv}. Similar to our method in encoding point cloud representation, FPConv~\cite{lin2020fpconv} aggregates features from point clouds onto 2D grids using a learned local flattening operation. Likewise, Point-PlaneNet~\cite{point-planenet} introduces a point cloud convolution operation called PlaneConv, which computes distances between points and aggregate them in a set of learned planes.
Nevertheless, all these methods do not consider the surface reconstruction task, which is the focus of our paper.

\textbf{Mesh representations:}
Meshes~\cite{gkioxari2019mesh,papier,human} emphasize topological relations by constructing vertices and faces, but mesh-based methods suffer from generating either shapes with only simple topology or self-intersecting meshes.

\textbf{Implicit neural representations:}
Recently, implicit occupancy~\cite{mescheder2018occupancy} and signed distance field~\cite{park2019deepsdf} have been exploited for 3D reconstruction. In contrast to the aforementioned explicit representations, implicit representation can model shapes in a continuous manner. Therefore, better detail preservation and more complicated shape topologies can be obtained from the implicit representation. Many recent works explore various applications, e.g., learning the implicit representation only from 2D observations~\cite{liu2019learning, liu2020dist,niemeyer2020differentiable}, encoding texture information~\cite{mildenhall2020nerf, oechsle2019texture}, or learning gradient fields~\cite{cai2020learning}. Unfortunately, all these methods are still limited to the reconstruction of single objects or small scenes with restricted complexity and struggle to generalize to scenes outside of the training distribution.

The notable exception is Peng~\etal~\cite{peng2020convolutional}. They propose an architecture that enables large-scale 3D scene reconstruction by training on synthetic indoor scene dataset and testing its generalization to larger scenes such as ScanNet \cite{dai2017scannet} and MatterPort3D \cite{chang2017matterport3d}.
Specifically, given a point cloud, this method projects point-wise features onto the canonical planes or volume grids and then use U-Net~\cite{unet} to aggregate both local and global information. In this way, the inductive biases are effectively exploited. However, considering only canonical planes may cause performance loss when some object parts do not align well with the canonical directions (\eg, a wired lamp with complicated geometry, see the first row of Fig.~\ref{fig:object}). Therefore, we propose to learn planes with a network. With such learned dynamic planes, our system shows better 3D reconstruction quality.

\section{Method}\label{sec:method}

Our goal is to reconstruct 3D scenes with fine details from noisy point clouds. To this end, we first encode the input point clouds into 2D feature planes, whose parameters are predicted by a fully-connected network. These feature planes are then processed using convolutional networks and decoded into occupancy probabilities via another shallow fully-connected network. Fig.~\ref{fig:method} illustrates the overall workflow of the proposed method.

\subsection{Encoder}

\begin{figure}[!ht]
\centering
\includegraphics[width=0.48\textwidth]{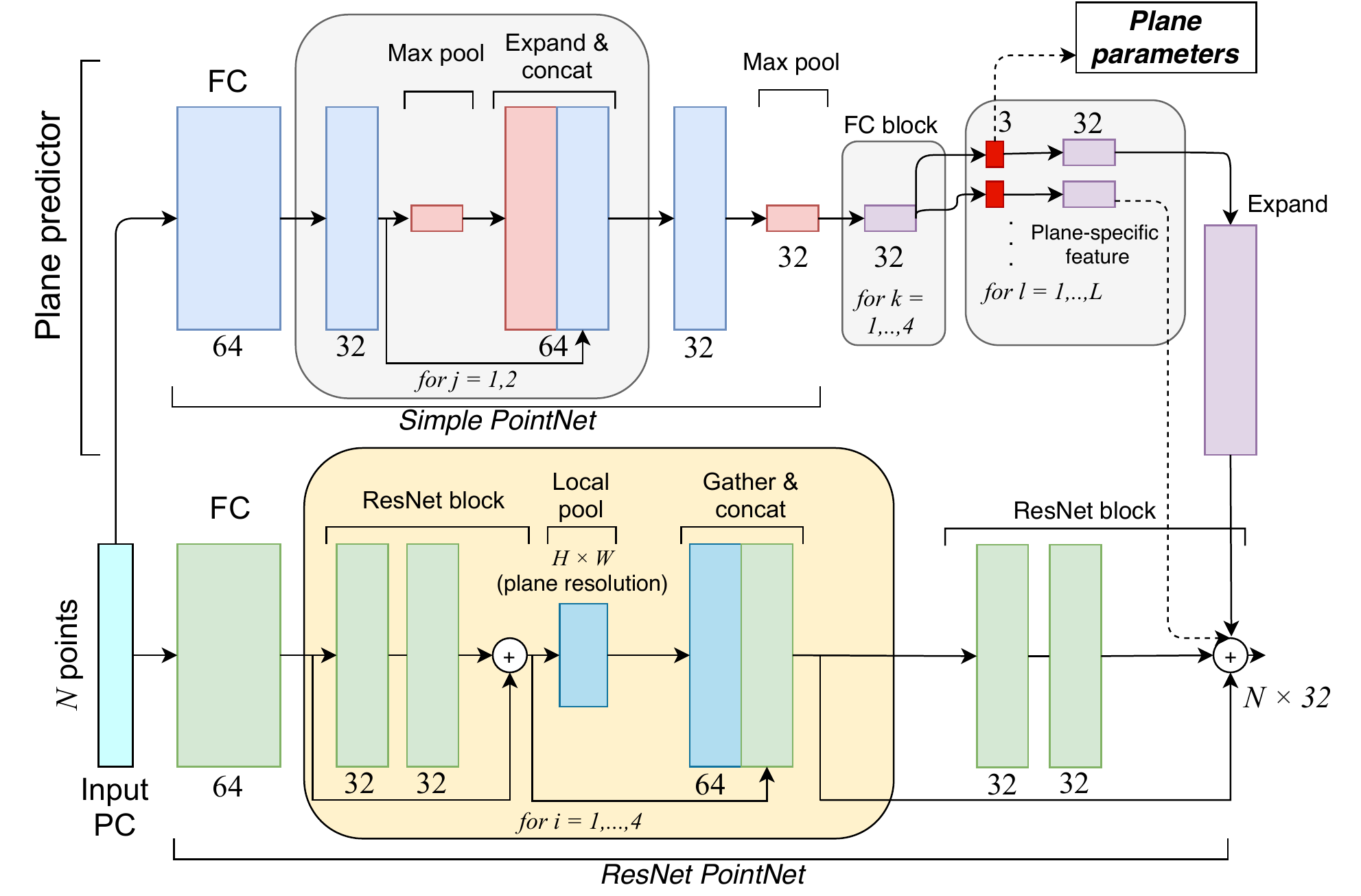}
\caption{\textbf{Encoder architecture.} We use a ResNet PointNet~\cite{mescheder2018occupancy,peng2020convolutional} to extract the per-point features. On top of it, we add a plane predictor network to predict the dynamic plane parameters, which consists of a simple PointNet~\cite{pointnet}.}
\label{fig:encoder}
\end{figure}

The architecture of our encoder is illustrated in Fig.~\ref{fig:encoder}. We describe each part as follows.

\boldparagraph{Point cloud encoding}
Given a noisy point cloud, we first form a feature embedding for every point with ResNet PointNet~\cite{mescheder2018occupancy}, in which we perform local pooling according to the predefined plane resolution~\cite{peng2020convolutional}.
We are applying a rather simple network here for the proof of concept, but other advanced feature extractors, e.g., PointNet++~\cite{pointnet++} or Tangent Convolution~\cite{tatarchenko2018tangent}, can also be used. 
 
\boldparagraph{Dynamic plane prediction}
Having the point-wise features, we can then construct the planar features.
Mathematically, a plane is defined by a normal vector $\mathbf{n}=(a, b, c)$ and a point $(x, y, z)$ which a plane passes through~\cite{point-planenet}: $ax + by + cz = d$.
Peng~\etal~\cite{peng2020convolutional} simply project features onto canonical planes, i.e., 3 planes aligned with the axes of the coordinate frame, $x = 0, y = 0, z =0$. 
Unlike~\cite{peng2020convolutional}, we introduce another shallow fully-connected network to regress the plane parameter $(a, b, c)$.
As illustrated in the upper branch of Fig.~\ref{fig:encoder}, we perform max pooling globally on all points in the point cloud because a global context is needed to search for the proposal of the best possible planes.
Since different input point clouds might be predicted with different planes, we call this process dynamic plane prediction.
Note that we directly set the intercept of the plane $d$ to 0 because the shifts along the normal direction do not change the feature projection process.

After the prediction of plane parameters, we pass it through one layer of FC to obtain a feature for every dynamic plane.
This feature is expanded, matching the number of input point cloud and summed up with the last layer of ResNet PointNet with respect to the individual dynamic plane, and thus we call it the plane-specific feature.
Our main intention is to allow backpropagation into the plane predictor network, but we also empirically find that this individual summation operation improves the reconstruction quality. One possible reason is that it allows the networks to learn varying \emph{emphasis} over the feature dimension of the last layer of ResNet PointNet with respect to the individual dynamic plane. 

Once having the predicted plane parameters, we project the summed-up features onto the dynamic planes with a defined size of $H \times W$ grids and apply max pooling for the features falling into the same grid cell.

\begin{figure}[!t]
\centering
\includegraphics[width=0.48\textwidth]{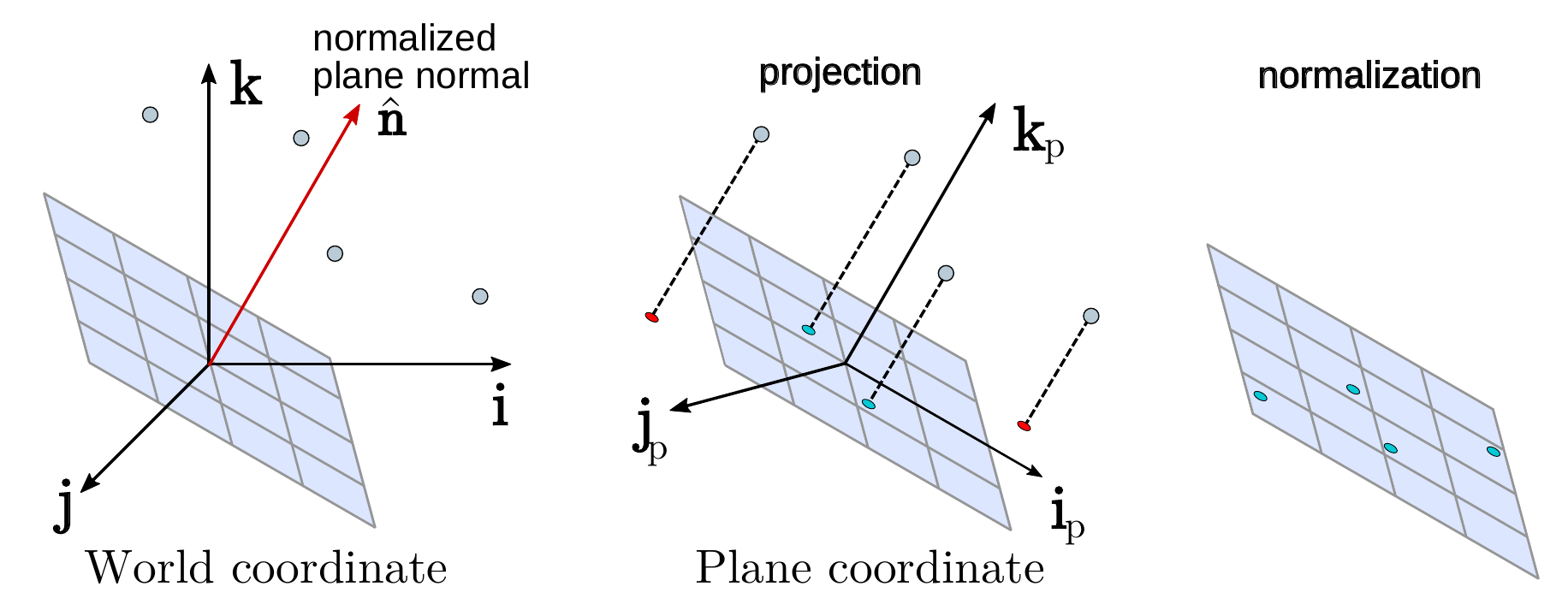}
\caption{\textbf{Planar projection.} To obtain the projected feature plane, we sequentially apply basis change, orthographic projection, and normalization.}
\label{fig:projection}
\end{figure}

\boldparagraph{Planar projection} In order to project the encoded features to the dynamic planes, whose normals can point to any directions, we sequentially apply basis change, orthographic projection, and normalization to always keep them inside $H \times W$ grids. Denoting the three basis vectors of canonical axes, $\mathbf{i}$, $\mathbf{j}$, $\mathbf{k}$, where $\mathbf{k}$ is the basis vector of the ground plane and $\mathbf{n}$ is the learned plane normal, those operations are detailed as follows and illustrated in Fig. \ref{fig:projection}.

To perform basis change, we normalize $\mathbf{n}$ into a unit vector $\hat{\mathbf{n}}$ and obtain the rotation matrix $R$ that aligns $\mathbf{k}$ with $\hat{\mathbf{n}}$. Let $\mathbf{v} =\begin{bmatrix}
v_1 & v_2 & v_3
\end{bmatrix}^{\top} = \mathbf{k} \times \hat{\mathbf{n}}$, the rotation matrix $R$ is defined as:

\begin{equation}
  R = I + [\mathbf{v}]_\times + [\mathbf{v}]^{2}_\times \frac{1- \mathbf{k} \cdot \hat{\mathbf{n}} }{\lVert \mathbf{v} \rVert^2},
\label{eq:rotation}
\end{equation}\\
where $ [\mathbf{v}]_\times$ is the skew matrix:
\begin{equation}
 [\mathbf{v}]_\times \coloneqq \begin{bmatrix}
0 & -v_3 & v_2 \\
v_3 & 0 & -v_1 \\
-v_2 & v_1 & 0
\end{bmatrix}
\label{eq:skew}
\end{equation}

With the rotation matrix $R$, we rotate the axes $\mathbf{i}$ and $\mathbf{j}$ to obtain $\mathbf{i}_{p}$ and $\mathbf{j}_{p}$. Now, the vectors $\mathbf{i}_{p}$, $\mathbf{j}_{p}$ and $\hat{\mathbf{n}}$ are orthogonal to each other, serving as the basis of the predicted plane coordinate system.

Next, we convert point coordinates from the world coordinate to the plane coordinate system and project the features orthographically to the predicted plane ("new ground plane"). 
However, as our dynamic plane can orient to any direction, the orthographic projection of a point far from the centroid of 3D space might fall outside the $H \times W$ grids. To ensure all possible points after orthographic projection are inside the grids, we divide the coordinates after projection by a normalization constant $c \geq 1$. To find $c$, we first convert $\mathbf{i}_{p}$ and $\mathbf{j}_{p}$ to be inside the positive octant by taking their absolute values $\mathbf{i}^+_{p}$ and $\mathbf{j}^+_{p}$. Next, we obtain orthogonal projections of the vector $\mathbf{1} = [1,1,1]^\top$ to $\mathbf{i}^+_{p}$ and $\mathbf{j}^+_{p}$:

\begin{equation}
\mathbf{a}_{\mathbf{i}} =  \frac{\mathbf{1}\cdot\mathbf{i}^+_{p}}{\mathbf{i}^+_{p} \cdot\mathbf{i}^+_{p}}\mathbf{i}^+_{p}, \;\;\;\;\;\;
\mathbf{a}_{\mathbf{j}} = \frac{\mathbf{1}\cdot\mathbf{j}^+_{p}}{\mathbf{j}^+_{p} \cdot\mathbf{j}^+_{p}}\mathbf{j}^+_{p}
\end{equation}
Subsequently, we set $c$ to be the maximum value between the lengths of these two projected vectors, $c =\max(|\mathbf{a}_{\mathbf{i}}|, |\mathbf{a}_{\mathbf{j}}|)$.
The point coordinates under the plane coordinate system are divided by $c$ so that all points lie inside the dynamic plane, where the point features are stored.

Once constructing the projected feature planes, we process them using U-Net \cite{unet} with shared weights for every plane. The final planar features have a dimension of $H \times W \times D$, where $D$ is the predetermined hidden dimension. 

\subsection{Decoder} \label{sec:decoder}

The goal of the decoder is to obtain the occupancy prediction of any point $\mathbf{p} \in \mathbb{R}^3$ given the aggregated planar features. Similar to how we project features in the encoder, we project $\mathbf{p}$ onto all dynamic planes. Next, we query the feature
through bilinear interpolation of the planar features encoded at the four neighboring plane grids.

\boldparagraph{Occupancy prediction} Given an input point cloud $\mathbf{x}$, we predict the occupancy of $\mathbf{p}$ based on the feature vector at point $\mathbf{p}$, denoted as $\psi(\mathbf{p}, \mathbf{x})$:
\begin{equation}
    f_{\theta}(\mathbf{p}, \psi(\mathbf{p}, \mathbf{x})) \rightarrow [0,1]
    \label{eq:occupancy_prediction}
\end{equation}

We use the same network as \cite{niemeyer2020differentiable}, which consists of 5 ResNet blocks with $\psi$ added to the input features of every block. The hidden dimension of all fully-connected layers is set to 32, which results in only $16,000$ parameters.

\subsection{Training and Inference}\label{sec:training}
During training, we apply binary cross-entropy loss between the occupancy prediction $ f_{\theta}(\mathbf{p}, \psi(\mathbf{p}, \mathbf{x}))$ and the true occupancy value of $\mathbf{p}$. During inference, we apply \emph{Multiresolution IsoSurface Extraction} (MISE) \cite{mescheder2018occupancy} to construct meshes.

\subsection{Positional Encoding}

The work of Mildenhall~\etal~\cite{mildenhall2020nerf} suggests that mapping input to higher dimension features using high-frequency functions before feeding into neural networks can result in the better fitting of data containing high-frequency variations. They introduce the positional encoding function:
\begin{equation}
\begin{split}
    \gamma(\mathbf{p}) = (\sin(2^0\pi \mathbf{p}), \cos(2^0\pi \mathbf{p}),\dots, \\
    \sin(2^{L-1}\pi \mathbf{p}), \cos(2^{L-1}\pi \mathbf{p}))
\end{split}
\label{eq:pos_encoding}
\end{equation}
where $L$ is the frequency band. While \cite{mildenhall2020nerf} verifies its effectiveness for image rendering tasks, we show that its functionality also generalizes to 3D point cloud reconstruction, as seen in Table~\ref{tab:shapenet_result}.

Specifically, we apply the positional encoding for the input 3D coordinates. Setting $L$ to 10, we map the input point cloud $\mathbf{x}$ and query points $\mathbf{p}$ from $\mathbb{R}^3$ to $\mathbb{R}^{60}$.

\section{Experiments}

To evaluate our method, we conduct two experiments on surface reconstruction from noisy point clouds. We perform \textbf{object-level reconstruction} using ShapeNet~\cite{chang2015shapenet} subset of Choy~\etal~\cite{choy20163d}, and \textbf{scene-level reconstruction} using synthetic indoor scene dataset from~\cite{peng2020convolutional}. 

\boldparagraph{Metrics} We follow the metrics used by \cite{peng2020convolutional}: \emph{Volumetric Intersection over Union (IoU)} measuring the matching volume of meshes intersection (higher is better), \emph{Chamfer-$L_1$} measuring the accuracy and completeness of the mesh surface (lower is better), \emph{Normal Consistency} measuring the accuracy and completeness of the mesh normals (higher is better, and \emph{F-score} measuring the harmonic mean of precision and recall between the reconstruction and ground truth (higher is better). The mathematical details are presented in the supplementary of \cite{peng2020convolutional}.

\boldparagraph{Implementation details} We use 32 as the hidden feature dimension for both encoder and decoder in all experiments, and Adam optimizer with a learning rate of $10^{-4}$. The depth of U-Net is chosen such that the receptive field
is equal to the size of the feature plane.
We choose a rather shallow fully-connected network as the plane predictor network. It has only around $13K$ parameters that are insignificant in size compared to the entire model, \eg, containing around $1.99M$ parameters for the model with 3 planes with a resolution of $64\times64$. The same depth of plane predictor network is used for the scene experiment. We run validation every 10,000 iterations and choose the best model based on the validation IoU.

\begin{table}[t]
\begin{center}
\resizebox{0.475\textwidth}{!}{%
\begin{tabular}{|l|ccccc|}
\hline
 & GPU & IoU & Chamfer- & Normal  & F-score \\
 & Memory & & $L_1$ & C. & \\
\hline\hline
\textbf{Without PE} & & & & & \\
ONet \cite{mescheder2018occupancy} & 7.7G & 0.761 & 0.087 & 0.891 & 0.785  \\
ConvONet (3C) \cite{peng2020convolutional} & 2.9G & 0.884 & 0.045 & 0.938 & 0.943 \\
Ours (3D) & 3.2G & 0.888 & 0.044 & 0.939 & 0.945 \\
Ours (5D) & 4.4G & 0.889 & 0.043 & 0.940 & 0.948 \\
Ours (7D) & 5.5G & 0.888 & 0.043 & 0.940 & 0.947  \\
Ours (3C + 2D) & 4.4G & 0.890 & 0.043 & 0.940 & 0.947  \\
Ours (3C + 4D) & 5.5G & 0.890 & 0.043 & 0.940 & 0.947  \\
\hline
\textbf{With PE} & & & & & \\
ConvONet (3C) \cite{peng2020convolutional} & 2.9G & 0.889 & 0.043 & 0.938 & 0.945  \\
Ours (3D) & 3.2G & 0.892 & 0.043 & 0.940 & 0.947 \\
Ours (5D) & 4.4G & 0.894 & \textbf{0.042} & \textbf{0.941} & 0.950  \\
Ours (7D) & 5.5G & \textbf{0.895} & \textbf{0.042} & \textbf{0.941} & 0.951  \\
Ours (3C + 2D) & 4.4G & 0.892 & 0.043 & \textbf{0.941} & 0.948 \\
Ours (3C + 4D) & 5.5G & 0.894 & \textbf{0.042} & \textbf{0.941} & 0.950 \\
\hline
\textbf{With PE + SL} & & & & & \\
Ours (3D) & 3.2G & 0.891 & 0.043 & 0.940 & 0.948  \\
Ours (5D) & 4.4G & 0.891 & 0.043 & \textbf{0.941} & 0.949  \\
Ours (7D) & 5.5G & \textbf{0.895} & \textbf{0.042} & \textbf{0.941} & \textbf{0.952} \\
\hline
\end{tabular}}\\
\footnotesize{PE = positional encoding. C = canonical planes. D = dynamic planes. SL = similarity loss.}

\end{center}
\caption{\textbf{Object-level 3D reconstruction from point clouds.} Results under all metrics are the mean for all 13 ShapeNet classes. The results for ONet \cite{mescheder2018occupancy} is taken from \cite{peng2020convolutional}.
Class-specific results can be found in supplementary.}
\label{tab:shapenet_result}
\end{table}

\subsection{Object-Level Reconstruction}

We first evaluate the task of single object reconstruction.
We sample 3000 points from the surface of ShapeNet objects and then apply Gaussian noise with zero mean and a standard deviation of 0.05. 
As for the query points (i.e., occupancy supervision), we follow~\cite{peng2020convolutional} and uniformly sample 2048 points. 
We use a plane resolution of $64^2$ and U-Net with a depth of 4.
The batch size during training is set to 32. 
All of the object-level models are trained until at least 900,000 iterations to ensure convergence.

\begin{figure*}[!t]
\begin{center}
\begin{tabular}{c|cc|cccc}
   & \multicolumn{2}{c|}{\textbf{Without PE}} & \multicolumn{4}{c}{\textbf{With PE}} \\
  GT & ONet & ConvOnet & ConvOnvet & Ours & Ours & Ours \\
  & \cite{mescheder2018occupancy} & (3C) \cite{peng2020convolutional} & (3C) \cite{peng2020convolutional} & (3D) & (5D) & (7D) \\
   & & & & & & \\
  \includegraphics[width=0.10\textwidth]{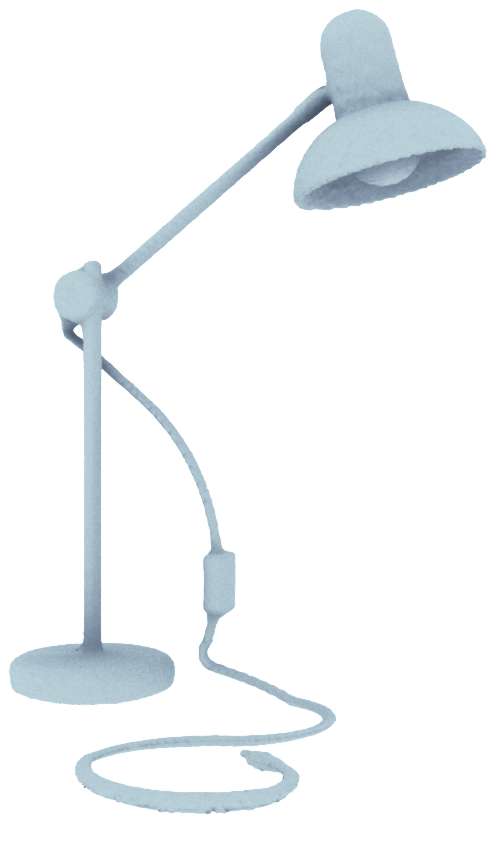} &   \includegraphics[width=0.10\textwidth]{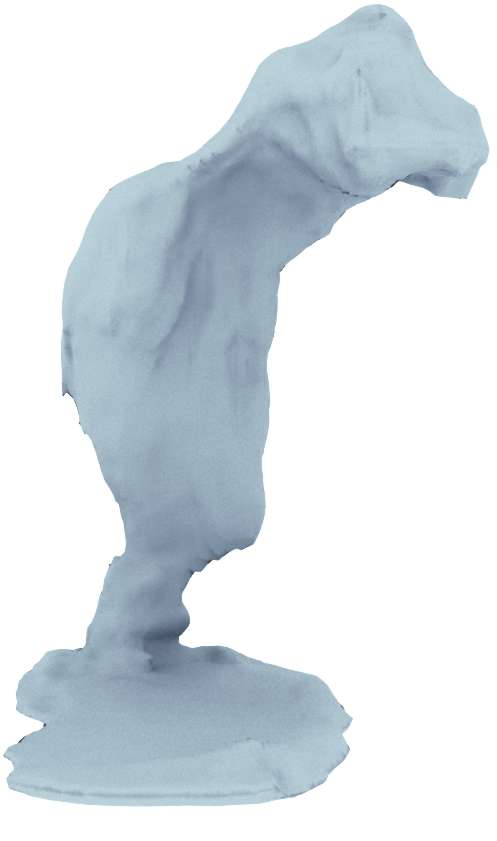} & 
  \includegraphics[width=0.10\textwidth]{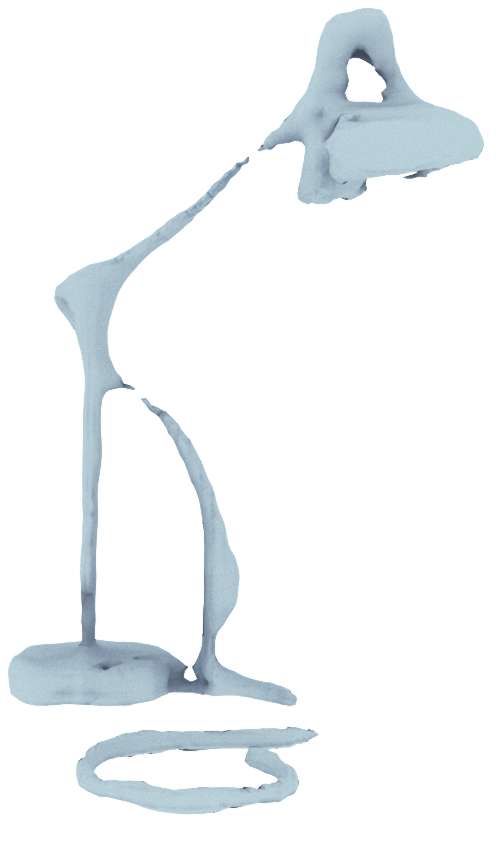} &
  \includegraphics[width=0.10\textwidth]{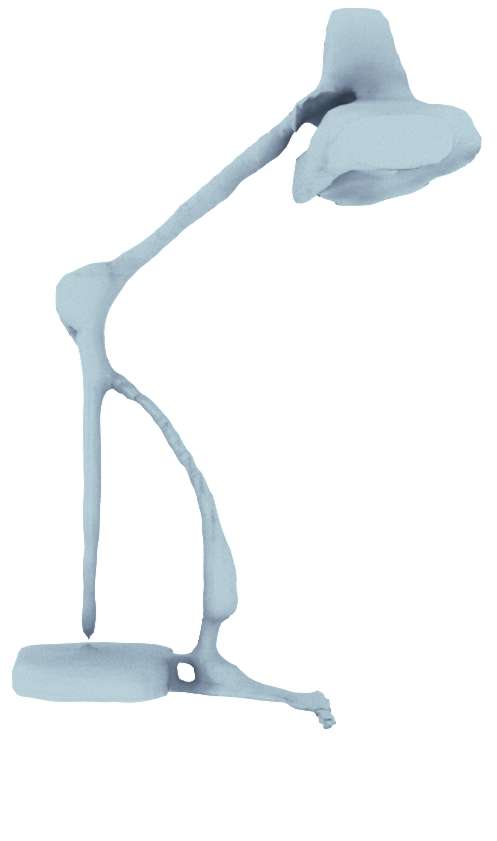} &
  \includegraphics[width=0.10\textwidth]{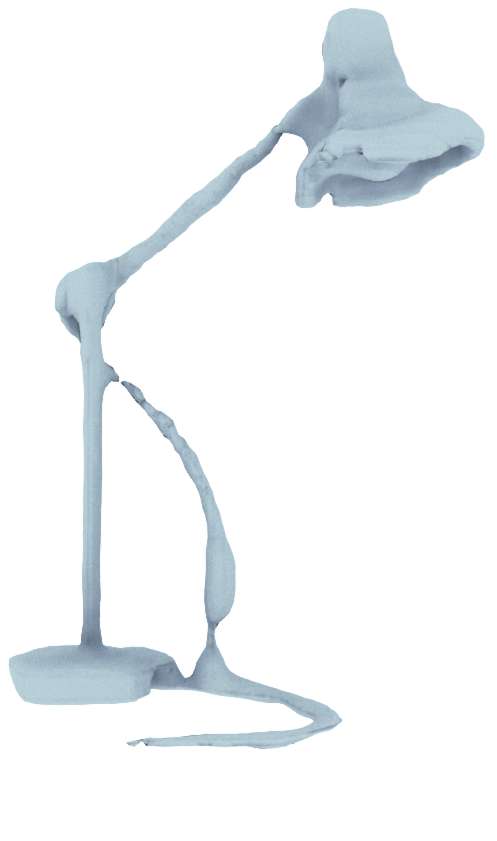} &
  \includegraphics[width=0.10\textwidth]{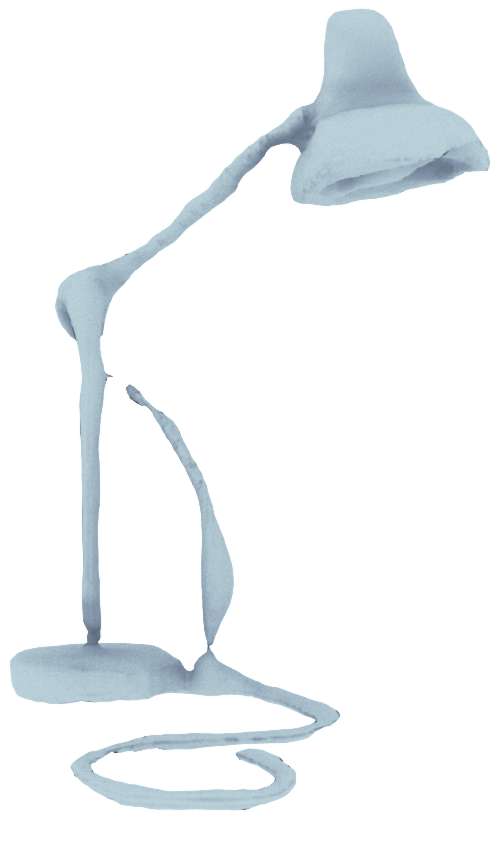} &
  \includegraphics[width=0.10\textwidth]{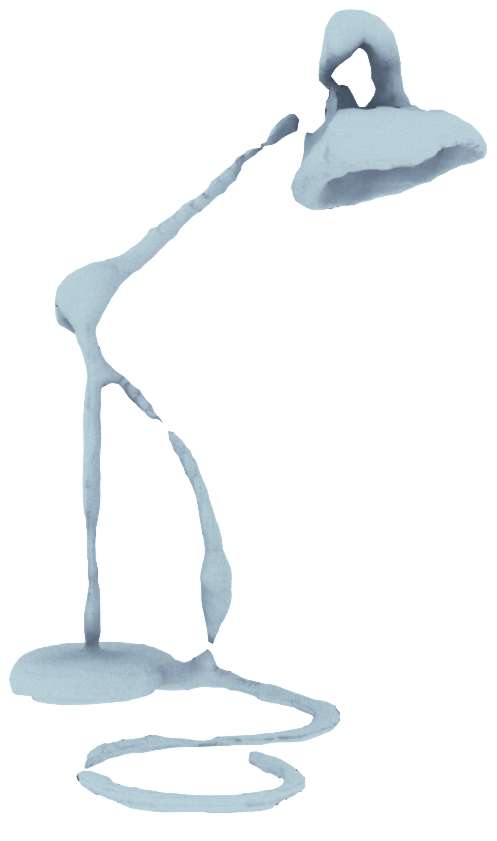}
  \\
  \includegraphics[width=0.10\textwidth]{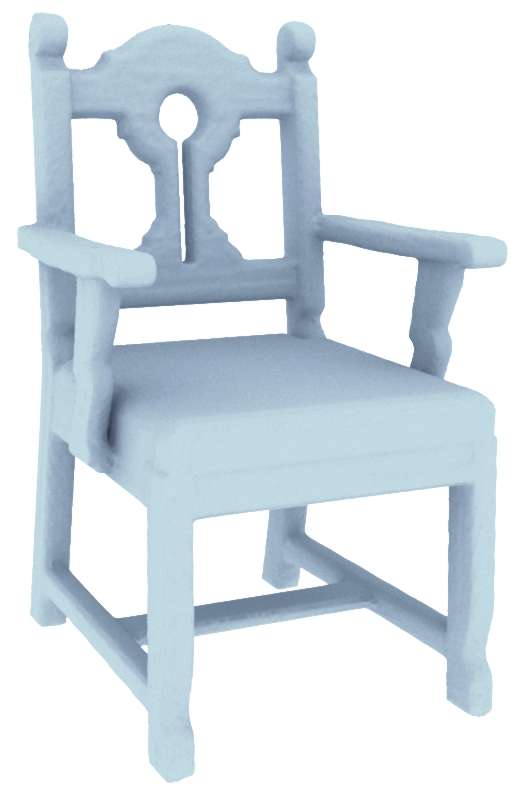} &   \includegraphics[width=0.10\textwidth]{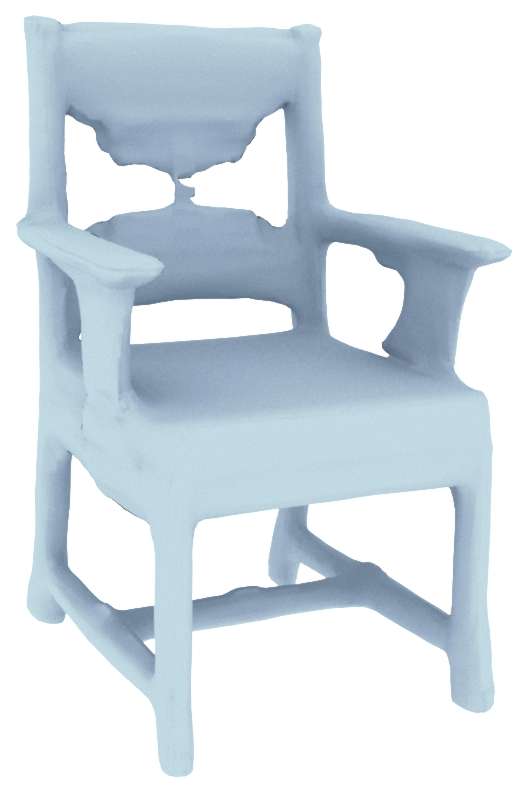} &
  \includegraphics[width=0.10\textwidth]{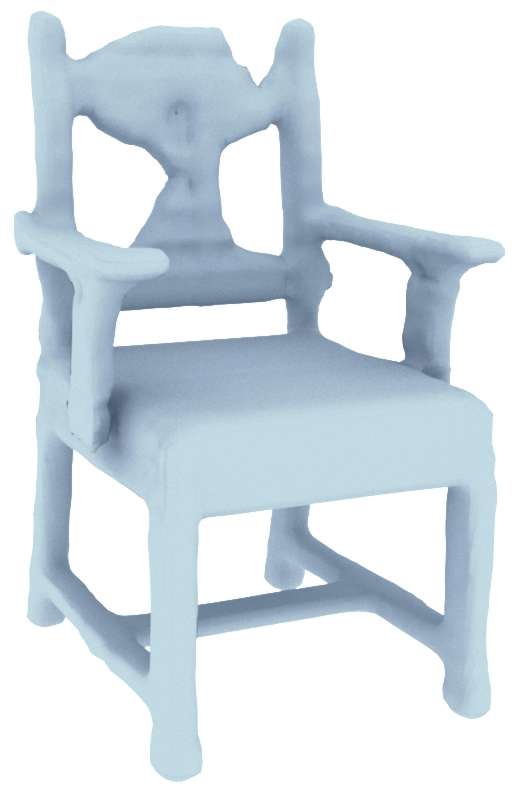} &
  \includegraphics[width=0.10\textwidth]{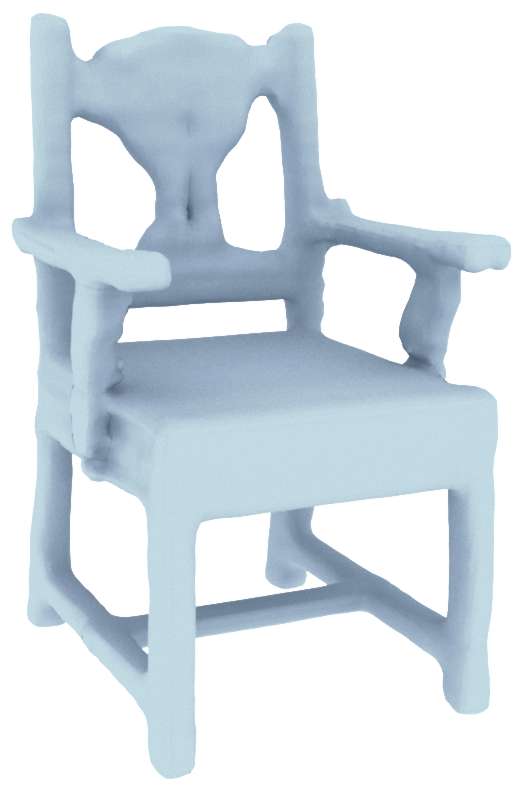} & 
  \includegraphics[width=0.10\textwidth]{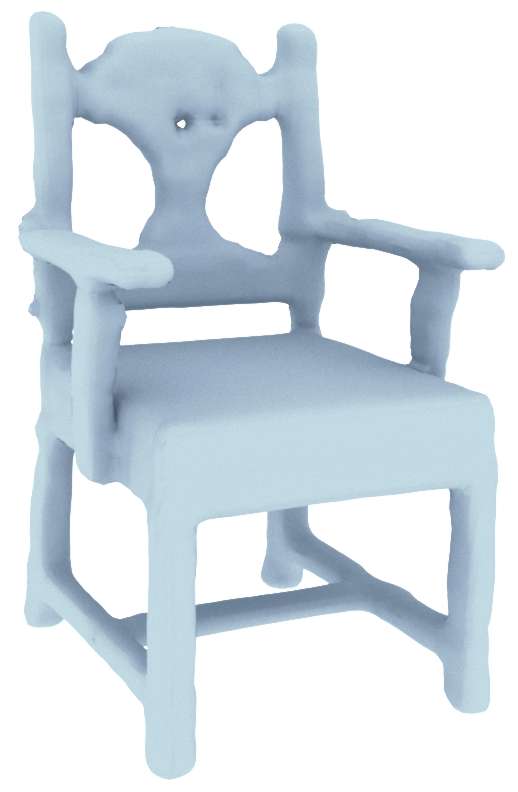} &
  \includegraphics[width=0.10\textwidth]{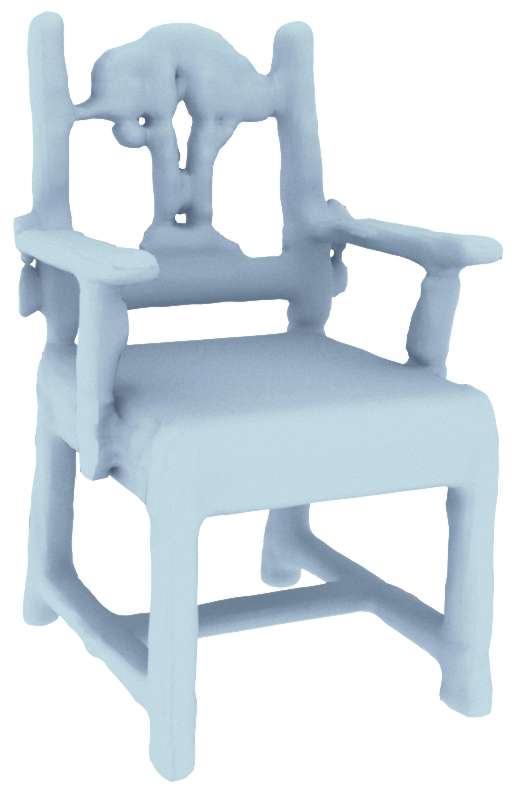} &
  \includegraphics[width=0.10\textwidth]{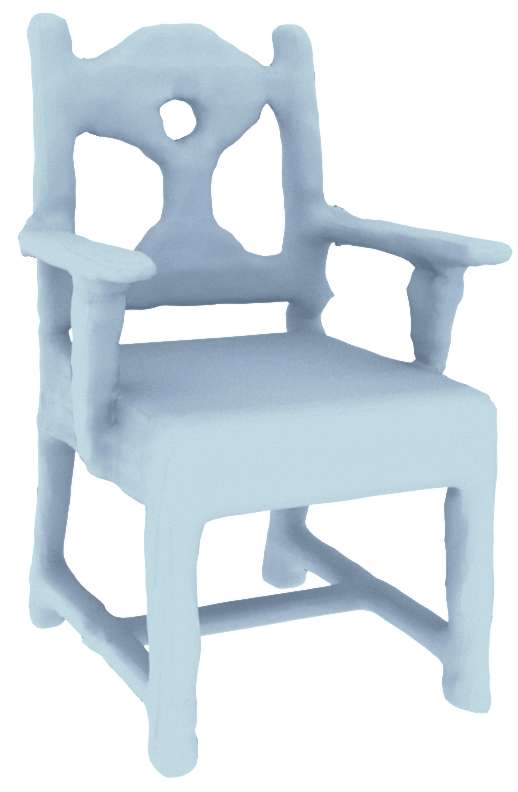} 
  \\
  \includegraphics[width=0.11\textwidth]{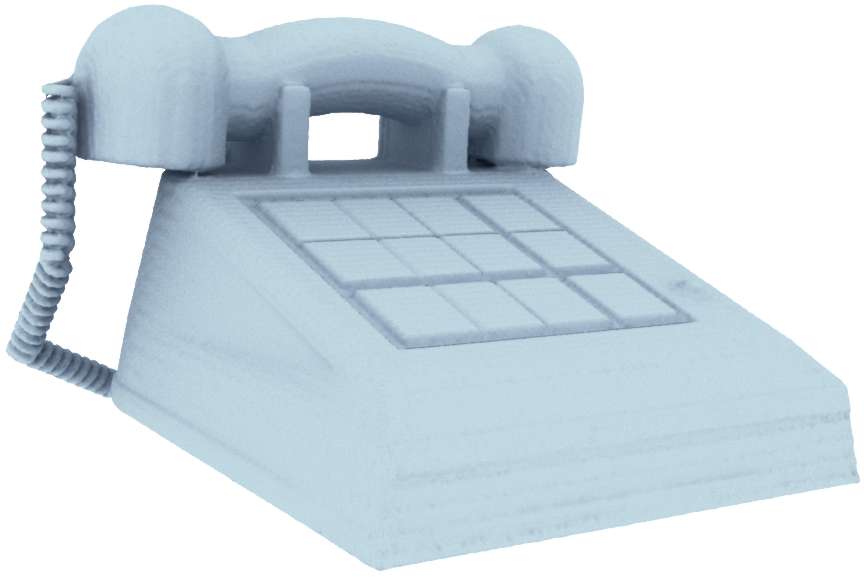} &
  \includegraphics[width=0.11\textwidth]{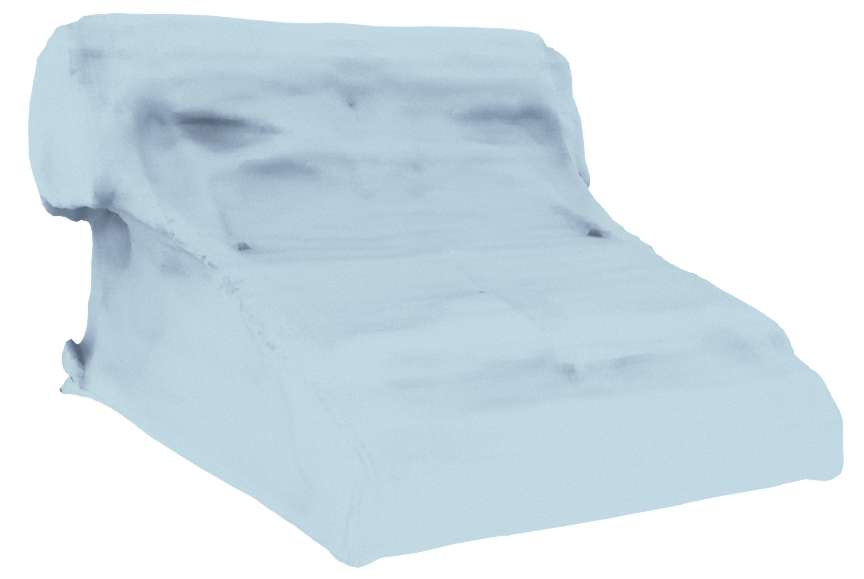} &   \includegraphics[width=0.11\textwidth]{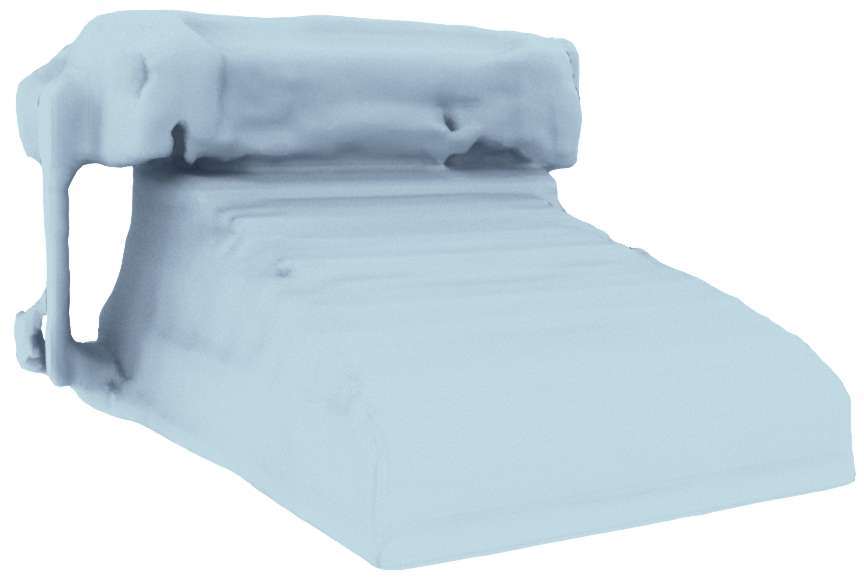} &
  \includegraphics[width=0.11\textwidth]{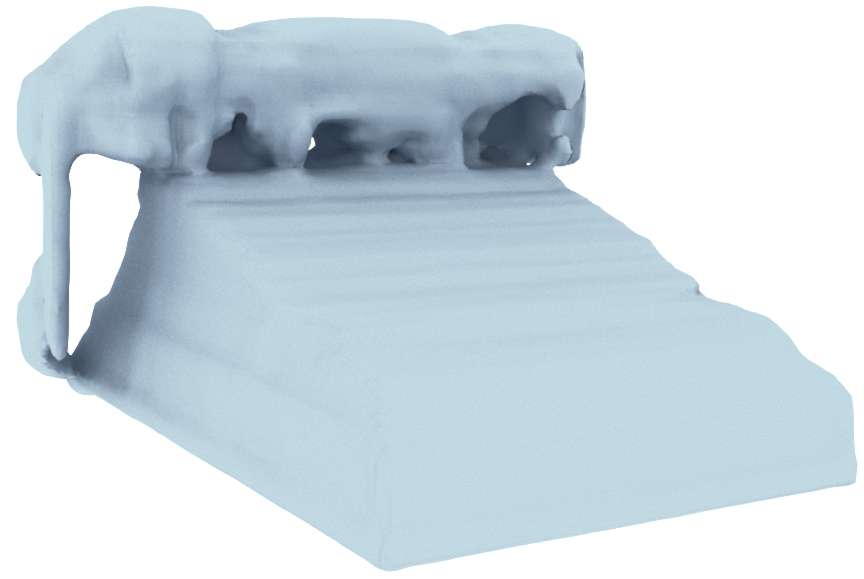} &
  \includegraphics[width=0.11\textwidth]{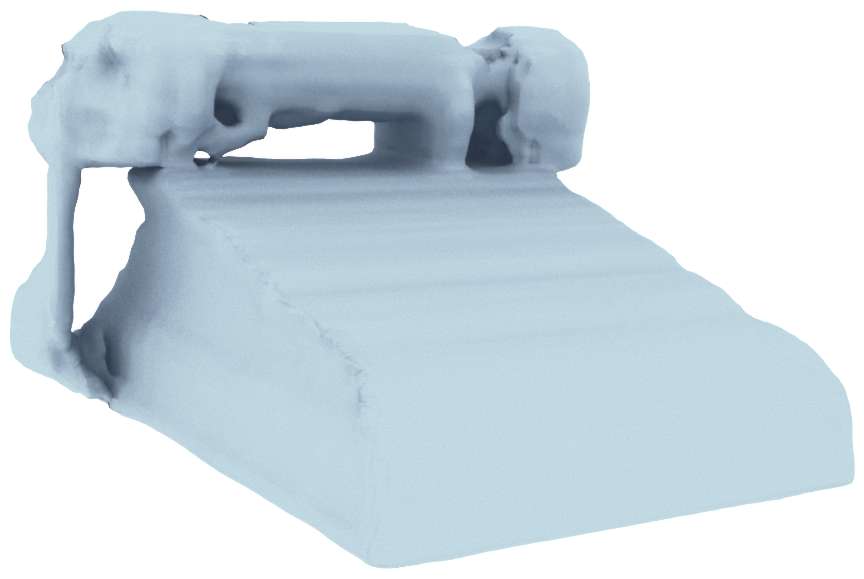} & 
  \includegraphics[width=0.11\textwidth]{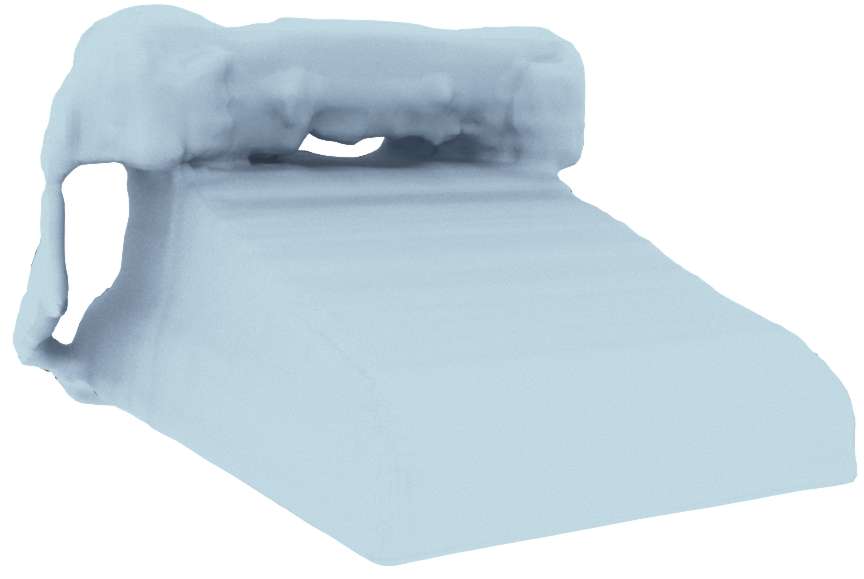} &
  \includegraphics[width=0.11\textwidth]{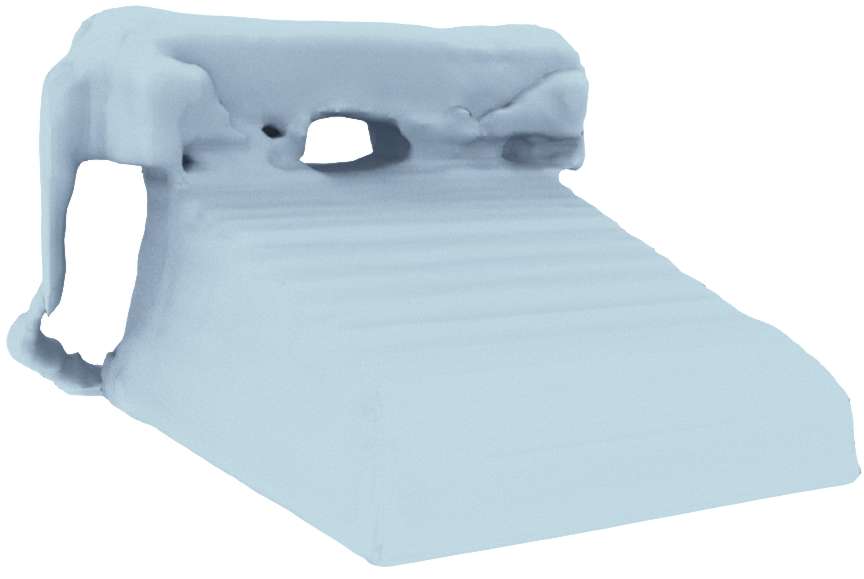} 
 
\end{tabular}
\end{center}
\caption{
\textbf{Object-level 3D reconstruction from point clouds.} Qualitative comparison
of our method to ONet~\cite{mescheder2018occupancy} and ConvONet~\cite{peng2020convolutional} on ShapeNet.}
\label{fig:object}
\end{figure*}

We run experiments with different combinations of canonical and dynamic planes. The results are summarized in Table \ref{tab:shapenet_result}. As we can notice, different variants of our method achieve state-of-the-art reconstruction accuracy on all metrics. Specifically, we outperform \cite{peng2020convolutional} while keeping the number of parameters at the same scale. We also observe progressive improvement when increasing the number of dynamic planes. Additionally, all results with positional encoding are better than without positional encoding. Moreover, as shown in our supplementary Section 2, we observe that adding positional encoding enables faster convergence. Qualitatively, the comparison against baselines is illustrated in Figure \ref{fig:object}. In general, the improvement from our models is more pronounced on the challenging classes and objects with intricate structures, such as thin components and holes. More elaborated results detailing per-category performance and more qualitative results are presented in the supplementary materials. 

\boldparagraph{Observation on plane distribution} 
Here, we discuss our observations on the distribution of the predicted dynamic planes.
In the case of 3 dynamic planes, our network predicts three canonical planes for all objects. 
This finding is interesting because it verifies the use of canonical planes in~\cite{peng2020convolutional}, and the canonical planes indeed describe various shapes most effectively, as ShapeNet objects are aligned along those axes. 
In the case of 5 and 7 dynamic planes, there are combinations of flipping sets of normals (\eg one normal pointing upward and the other downward). Such flipping sets of normals are equivalent to applying a horizontal flip on the projected encoded features. This observation is appealing because some recent works~\cite{wu2020unsupervised,xu2020ladybird} explicitly inject the object symmetry prior knowledge during training and show superior performance, while in our case, this symmetric property is implicitly encoded into our learned feature planes.
Indeed, many objects in ShapeNet are symmetric about a plane, \eg, most cars and airplanes are horizontally symmetric. We also conduct an ablation study evaluating the performance of ConvONet~\cite{peng2020convolutional} with 5 and 7 pre-defined static planes. The results of this ablation study further verify the superiority of our method. Details are in the supplementary materials.

\boldparagraph{Similarity loss} To test whether having diverse plane normals that are neither aligned nor flipping to each other can have a significant impact on the model performance, we try another variant where we restrict the learned plane normals to be diverging by adding a pairwise cosine similarity loss among plane normals, as defined below:

\begin{equation}
    \mathcal{L}_{similarity} = \frac{1}{M}\sum\limits_{\substack{i,j \\
                        i \neq j}}^{M}|(\cos(\theta_{i,j}))^{d}|
\label{eq:simloss}
\end{equation}\\
where $\theta_{i,j}$ is the angle between the pairwise plane normal pair, and $M$ is the total number of pairs. To ensure diverging plane normals, we set $d = 10$ so that the similarity loss starts penalizing when $\theta_{i,j}<45^{\circ}$ or $\theta_{i,j}>135^{\circ}$. With the additional similarity loss, the loss function is in the following form:

\begin{equation}
    \mathcal{L} = \mathcal{L}_{CE} + C \cdot \mathcal{L}_{similarity}
\label{eq:totalloss}
\end{equation}\\
where $\mathcal{L}_{CE}$ is the binary cross-entropy loss on the predicted occupancy in Eq.~\eqref{eq:occupancy_prediction}. In our experiments, we set $C$ to be $10 \times M$. The component from the similarity loss quickly converges to 0 when the predicted planes become diverse.

\begin{figure*}[!t]
\centering
\includegraphics[width=0.98\textwidth]{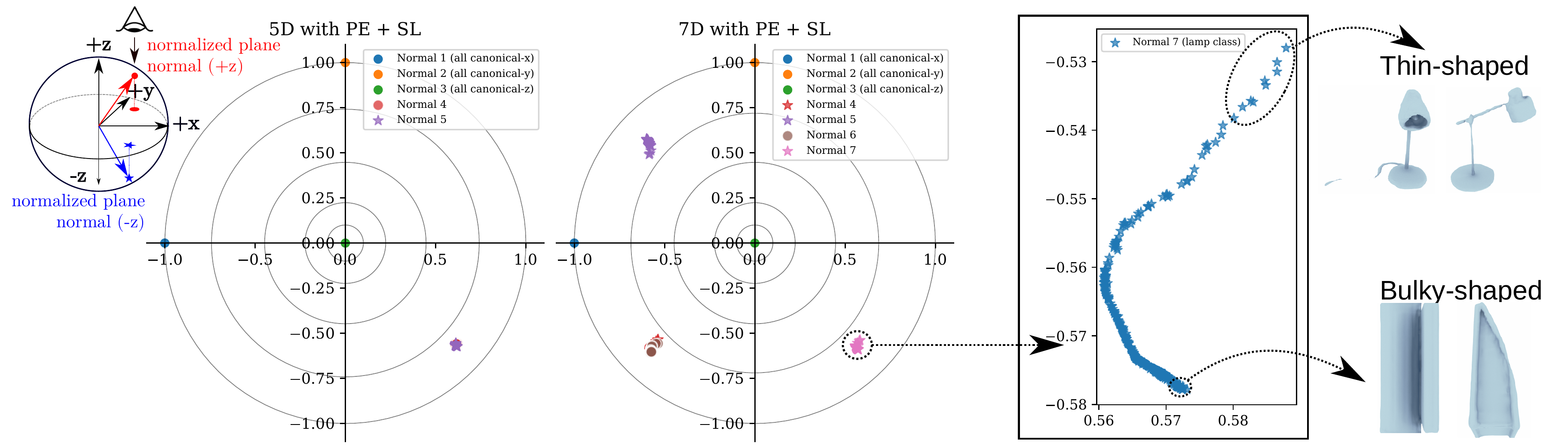}
\caption{\textbf{Plane normal distribution when trained with similarity loss.} We consider 5 and 7 dynamic plane models. In the three figures above, plane normals are normalized into unit lengths and projected as if viewed from the top. "$\bullet$" indicates the normal has $+z$ direction, while "$\star$" indicates $-z$ direction. \emph{Left:} 5 dynamic planes (all classes). \emph{Middle:} 7 dynamic planes (all classes). \emph{Right:} The distribution of one of the plane normals with slight variations of class "lamp" in 7 dynamic planes. It is observed that within this small variation, objects with different global structures favor different regions.}
\label{fig:plane_distribution}
\end{figure*}

With the similarity loss, diverging plane normals are observed. For 3 dynamic planes, the predicted planes are almost identical sets of planes in canonical axes. Adding more planes, \eg in 5 and 7 dynamic planes, gives predicted planes whose normals diverge from the canonical axes without any flipping or redundant set where slight variations between objects are observed. The plane distributions of the models with 5 and 7 planes trained with similarity loss (both with positional encoding) are illustrated in Fig.~\ref{fig:plane_distribution}. We observe that within the plane predictions with slight variations, objects having different global structures favor different regions, corroborating our networks' ability to learn planes that can vary based on the shape of an individual object. In terms of performance, however, we see little or no improvement compared to the unrestricted version. The results with the similarity loss are shown in Table \ref{tab:shapenet_result}.

\begin{figure}[!h]
\centering
\includegraphics[width=0.48\textwidth]{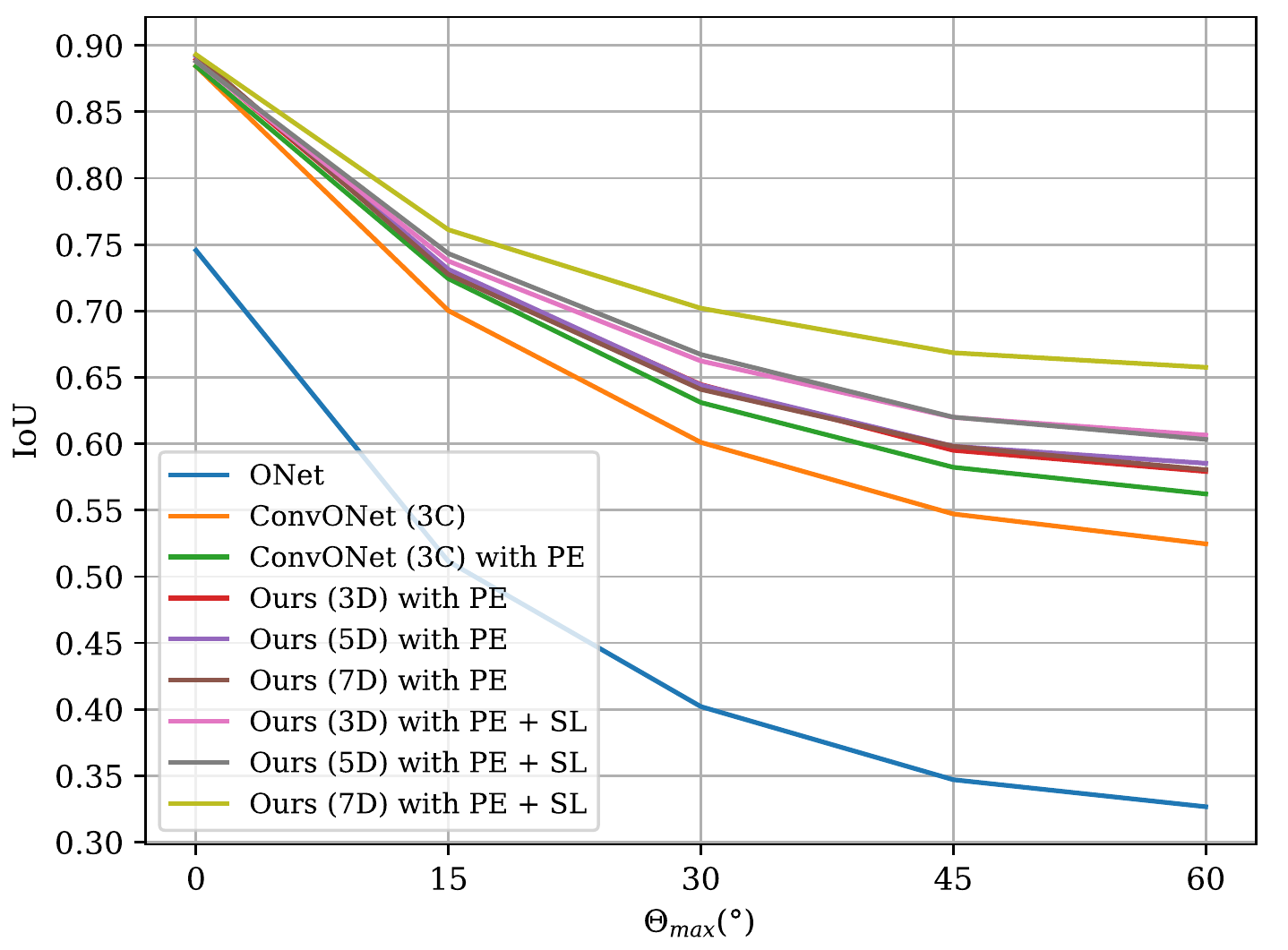}\\
\footnotesize{C = canonical planes. D = dynamic planes. PE = positional encoding. SL = similarity loss.}
\caption{\textbf{Rotation experiment on ShapeNet.} Comparison of IoU on ShapeNet test set rotated with angles uniformly sampled from $[0^{\circ}, \Theta_{max}]$ }
\label{fig:iou_drop}
\end{figure}

Interestingly, as shown in Fig. \ref{fig:iou_drop}, we see that the models trained with similarity loss have better generalization towards inputs with unseen orientations that have not been trained on, especially when using a high number of planes. We test the generalization towards different orientation by applying random rotation to the input in ShapeNet test set along $\mathbf{x}$, $\mathbf{y}$, and $\mathbf{z}$ axes with angles $\theta_x$, $\theta_y$, and $\theta_z$ drawn uniformly from $[0^{\circ}, \Theta_{max}]$ for each sample. Figure \ref{fig:iou_drop} shows the results of the experiments where the models are trained with all objects in canonical pose and tested on rotated poses. We can clearly notice the progressive drop of IoU when the test set is rotated with random angles up to $\Theta_{max}$. It can be seen that there is a considerable generalization improvement when using 7 dynamic planes trained with similarity loss. Comparing~\cite{peng2020convolutional} with or without positional encoding, we also see that positional encoding improves the generalization towards input in different orientations.

\begin{figure*}[ht!]
\begin{center}
\begin{tabular}{@{}c@{}@{}c@{}@{}c@{}@{}c@{}@{}c@{}@{}c@{}}

   \centered{\rotatebox[origin=c]{90}{GT}} & 
   &
  \centered{\includegraphics[width=0.15\textwidth]{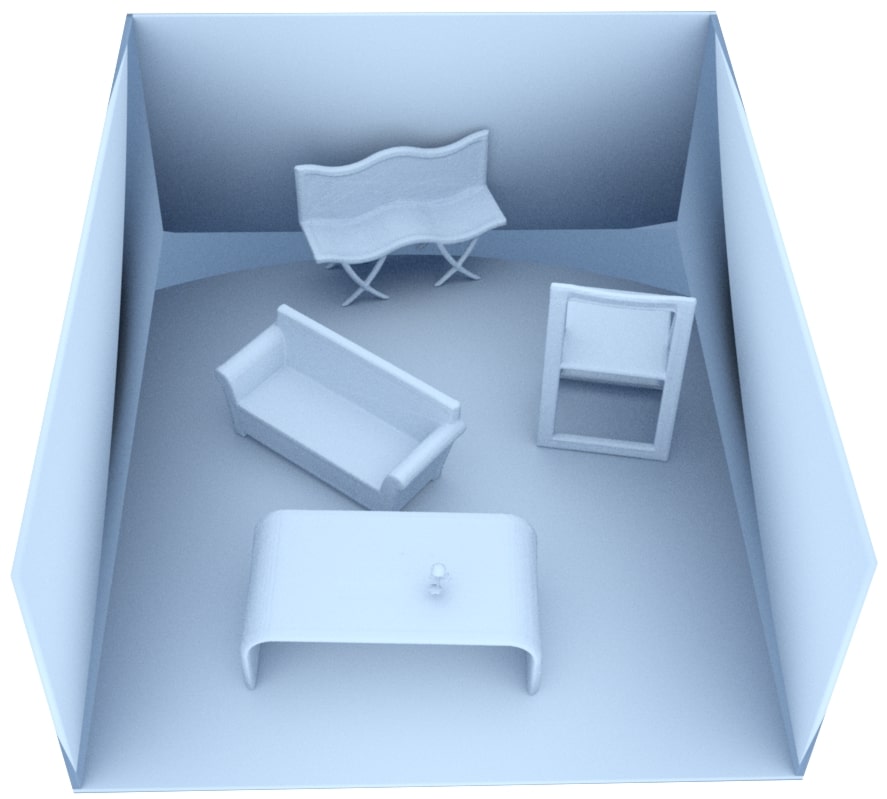}} &   \centered{\includegraphics[width=0.16\textwidth]{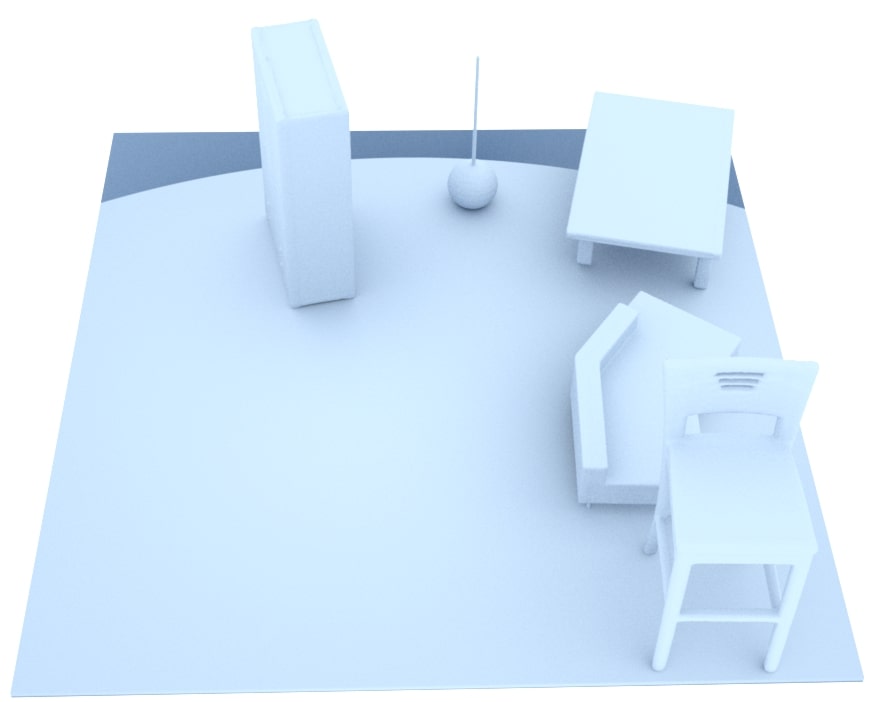}} &  
  \centered{\includegraphics[width=0.23\textwidth]{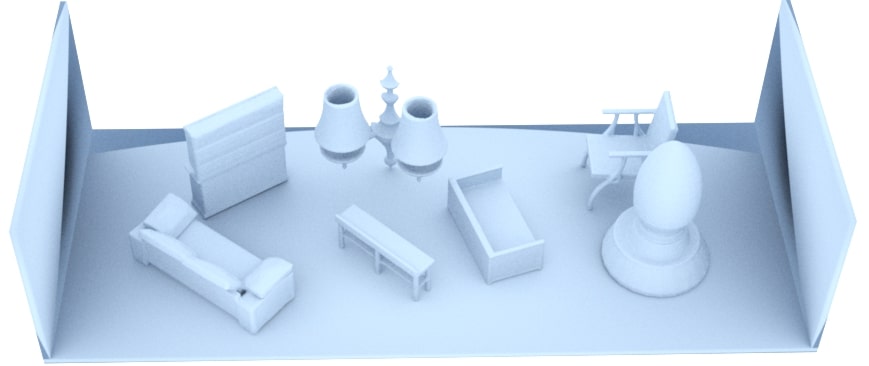}} &  \centered{\includegraphics[width=0.11\textwidth]{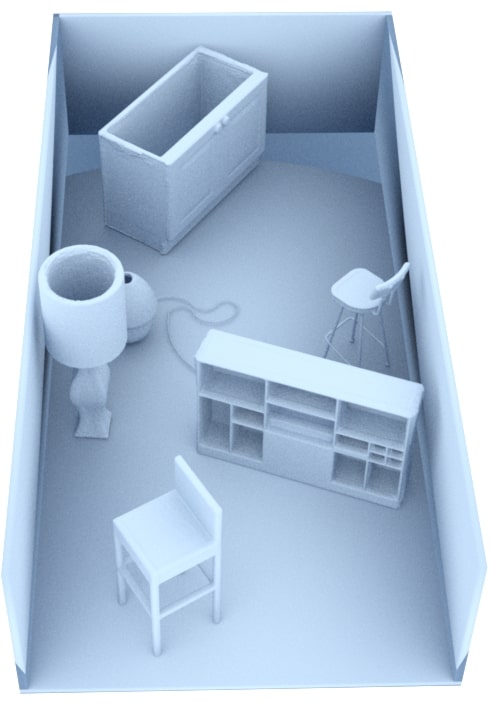}} \\
  \centered{\rotatebox[origin=c]{90}{ONet \cite{mescheder2018occupancy}}} &
   &
  \centered{\includegraphics[width=0.15\textwidth]{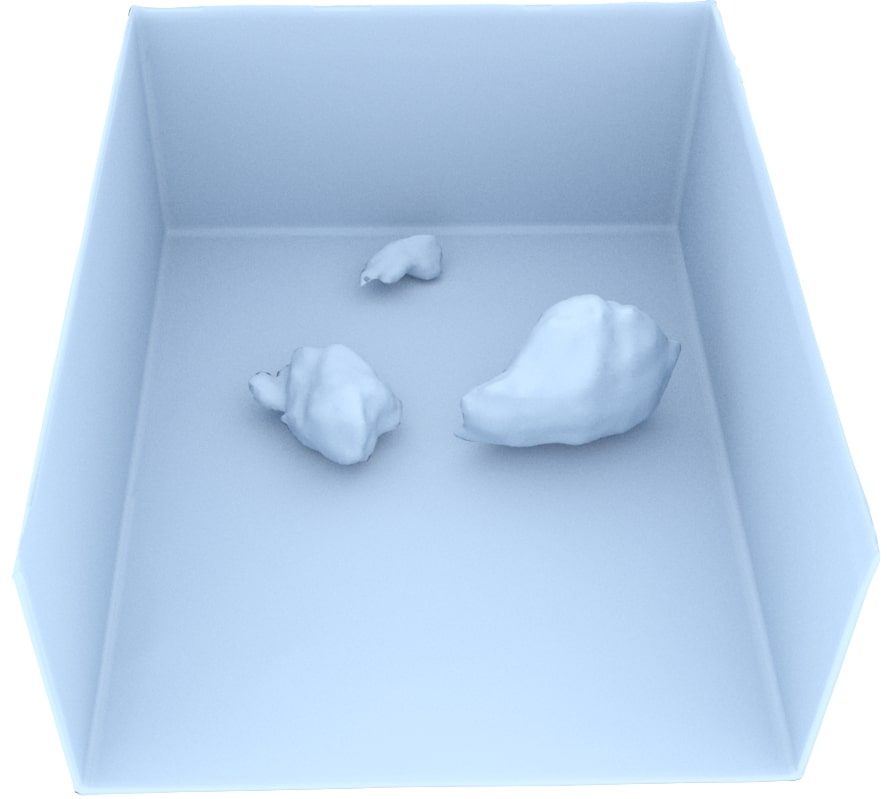}} &   \centered{\includegraphics[width=0.17\textwidth]{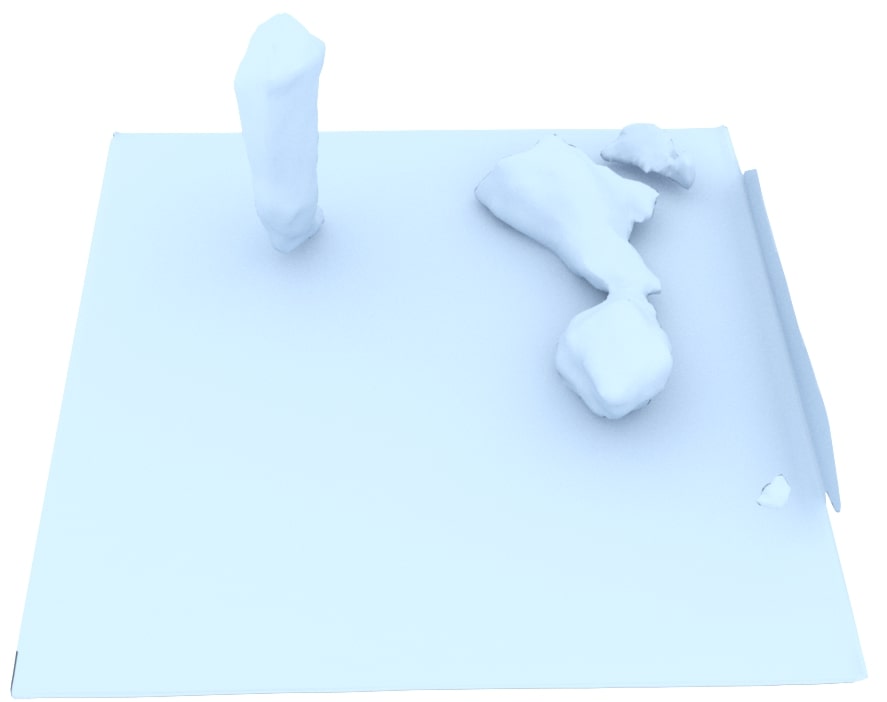}} &  
  \centered{\includegraphics[width=0.23\textwidth]{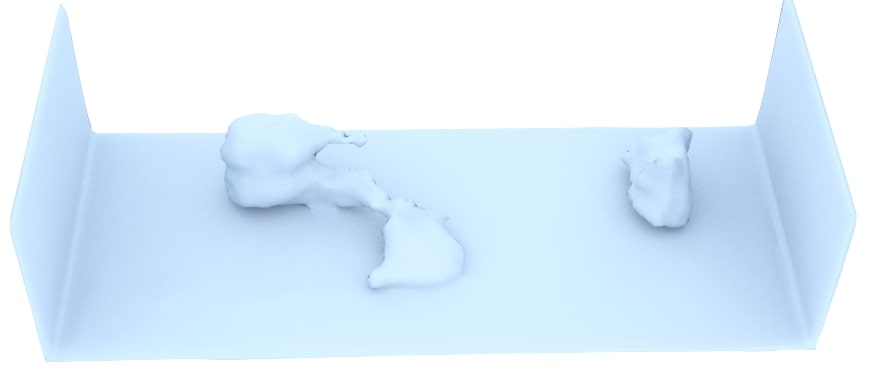}} &  \centered{\includegraphics[width=0.105\textwidth]{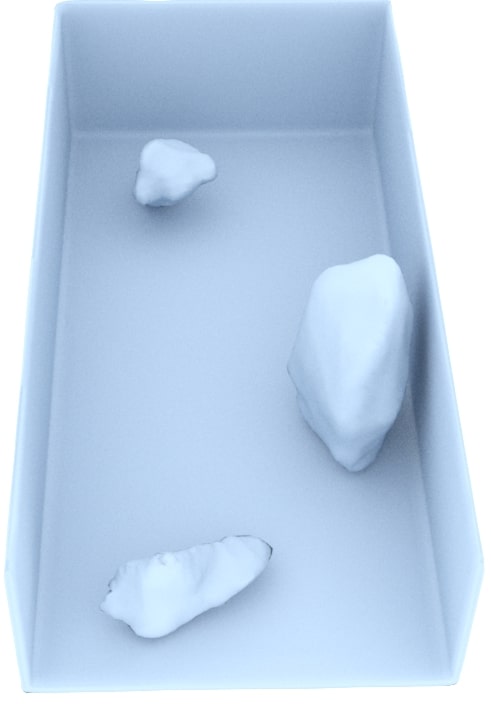}} \\
  \centered{\rotatebox[origin=c]{90}{ConvONet}} & 
  \rotatebox[origin=c]{90}{(3C) \cite{peng2020convolutional}} &
  \centered{\includegraphics[width=0.15\textwidth]{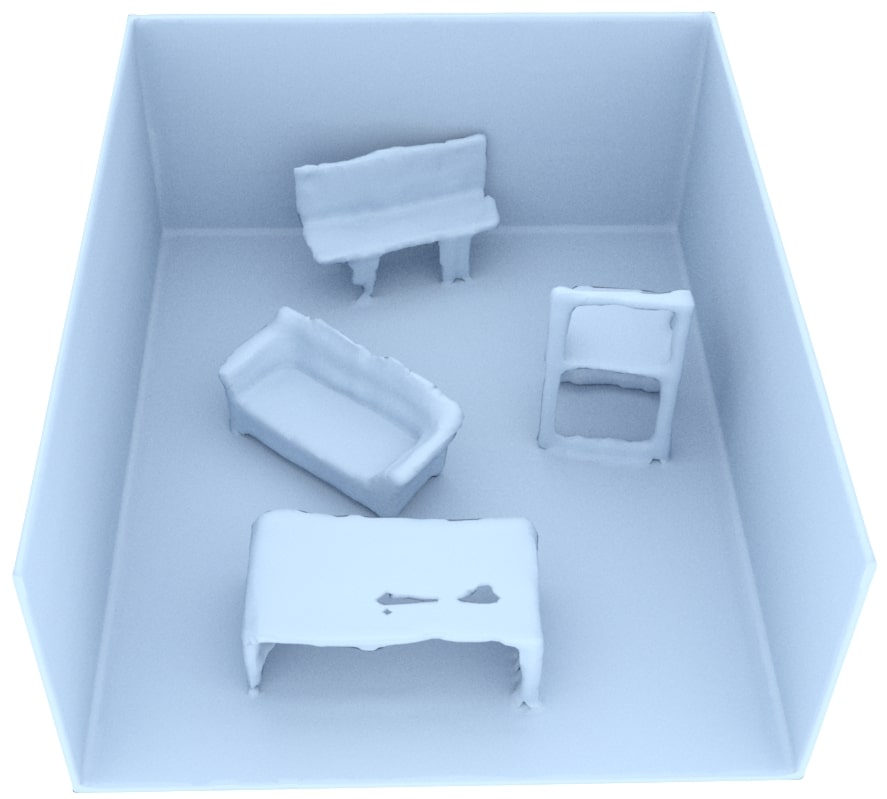}} &   \centered{\includegraphics[width=0.17\textwidth]{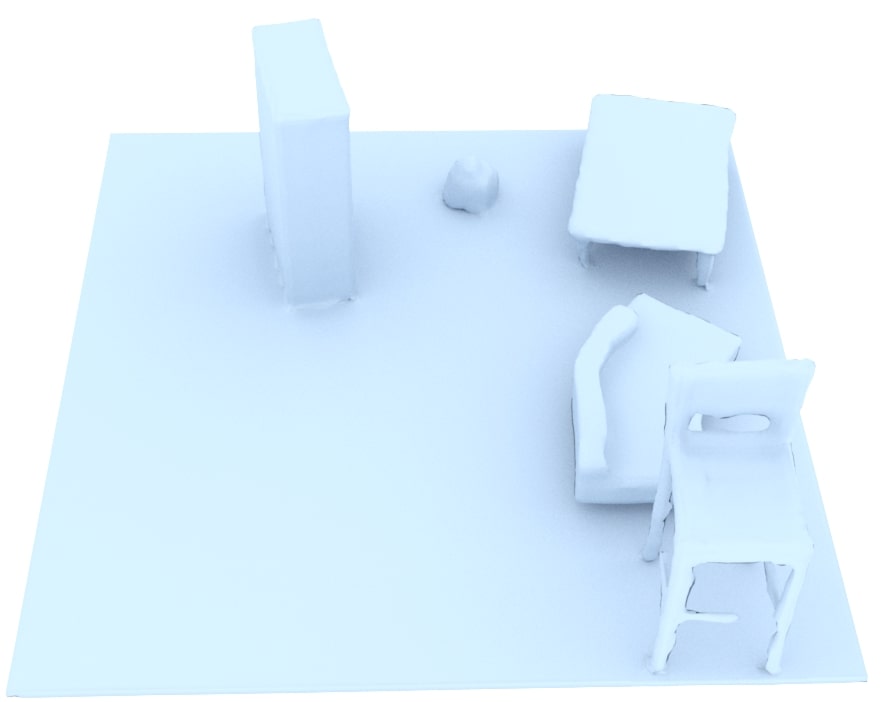}} &  
  \centered{\includegraphics[width=0.23\textwidth]{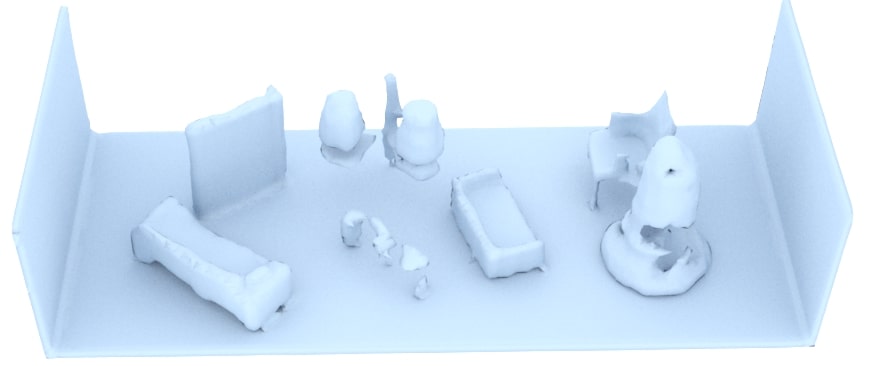}} &  \centered{\includegraphics[width=0.11\textwidth]{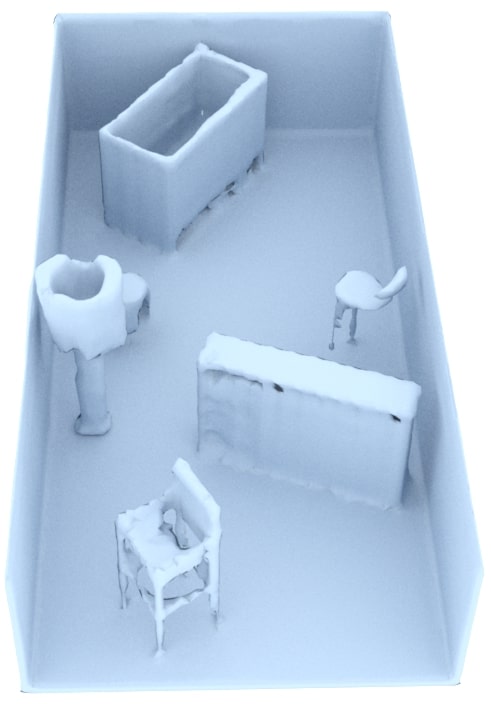}} \\
  \centered{\rotatebox[origin=c]{90}{ConvONet}} & 
  \rotatebox[origin=c]{90}{(3C + $32^3$) \cite{peng2020convolutional}} &
  \centered{\includegraphics[width=0.15\textwidth]{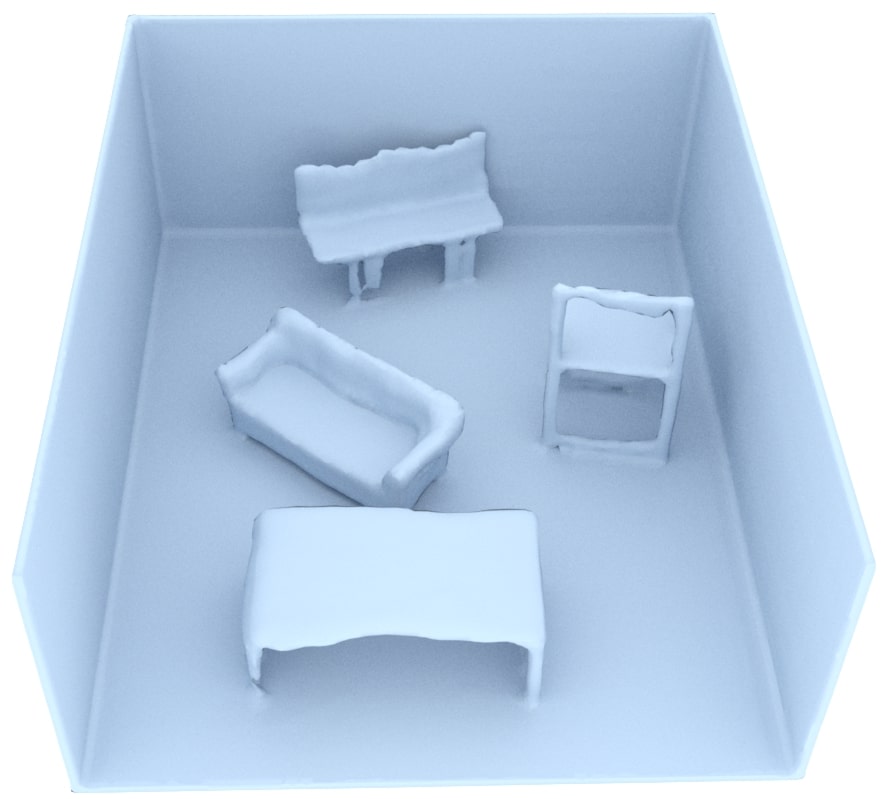}} &   \centered{\includegraphics[width=0.18\textwidth]{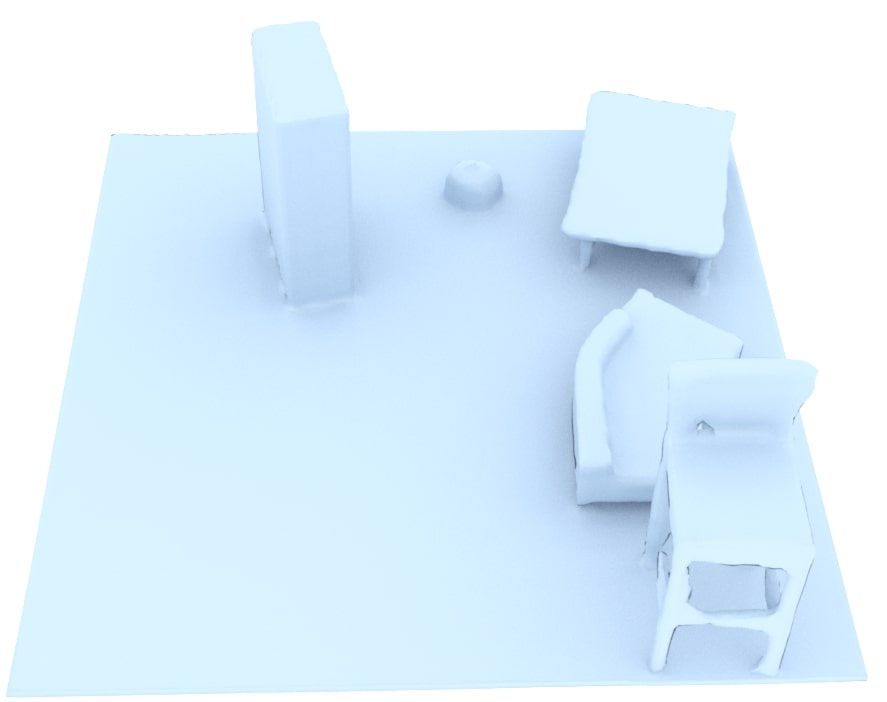}} &  
  \centered{\includegraphics[width=0.23\textwidth]{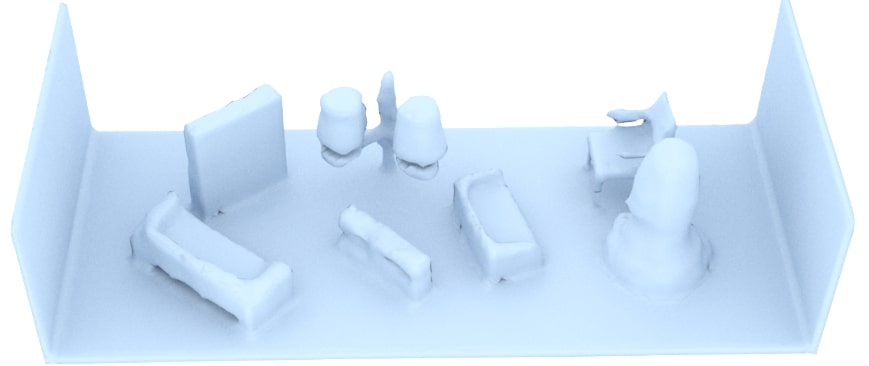}} &  \centered{\includegraphics[width=0.11\textwidth]{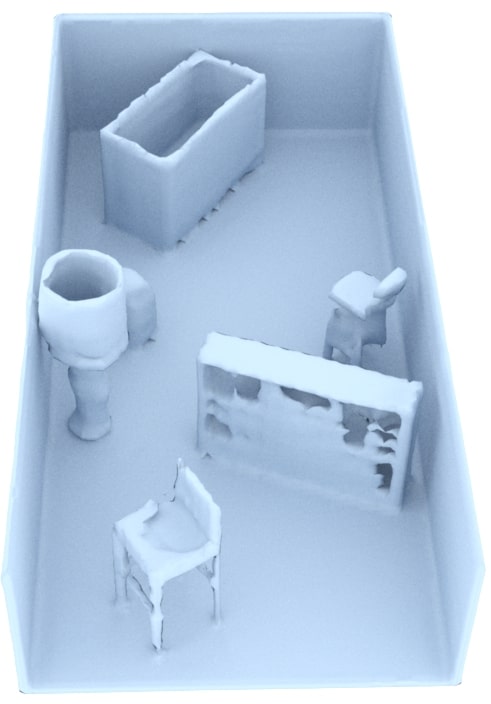}} \\
  \centered{\rotatebox[origin=c]{90}{Ours}} & 
  \rotatebox[origin=c]{90}{(3D)} &
  \centered{\includegraphics[width=0.15\textwidth]{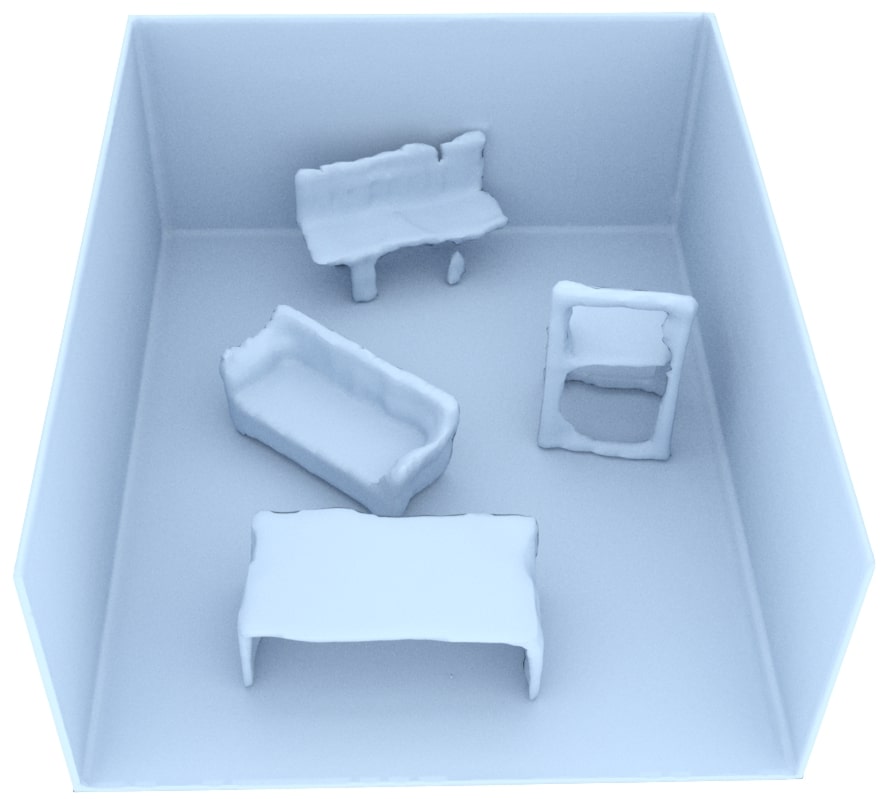}} &   \centered{\includegraphics[width=0.17\textwidth]{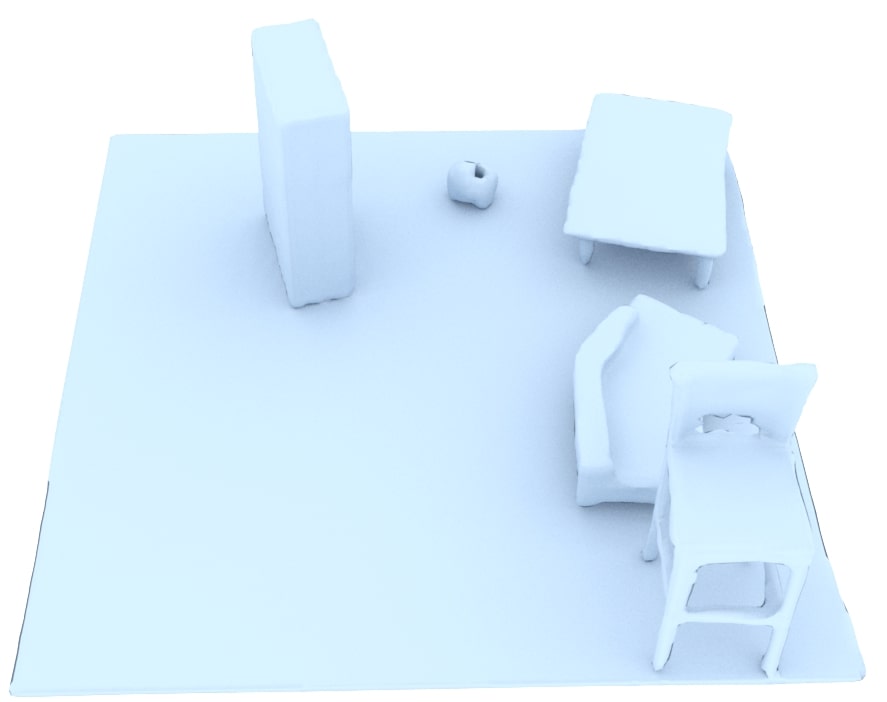}} &  
  \centered{\includegraphics[width=0.23\textwidth]{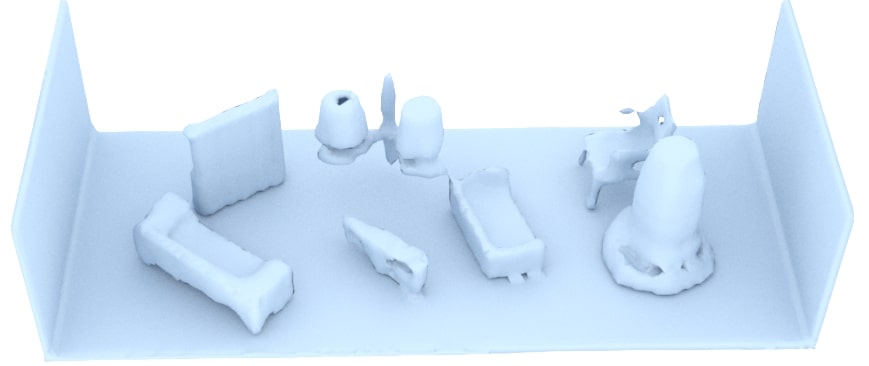}} &  \centered{\includegraphics[width=0.11\textwidth]{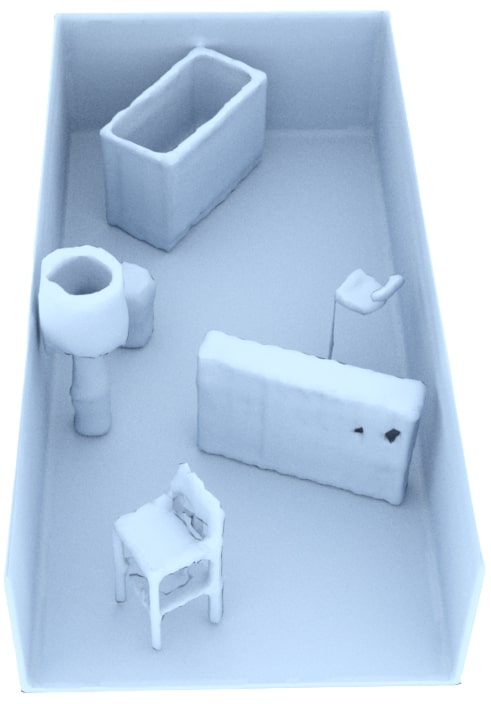}} \\
  \centered{\rotatebox[origin=c]{90}{Ours}} & 
  \rotatebox[origin=c]{90}{(3C+2D)} &
  \centered{\includegraphics[width=0.15\textwidth]{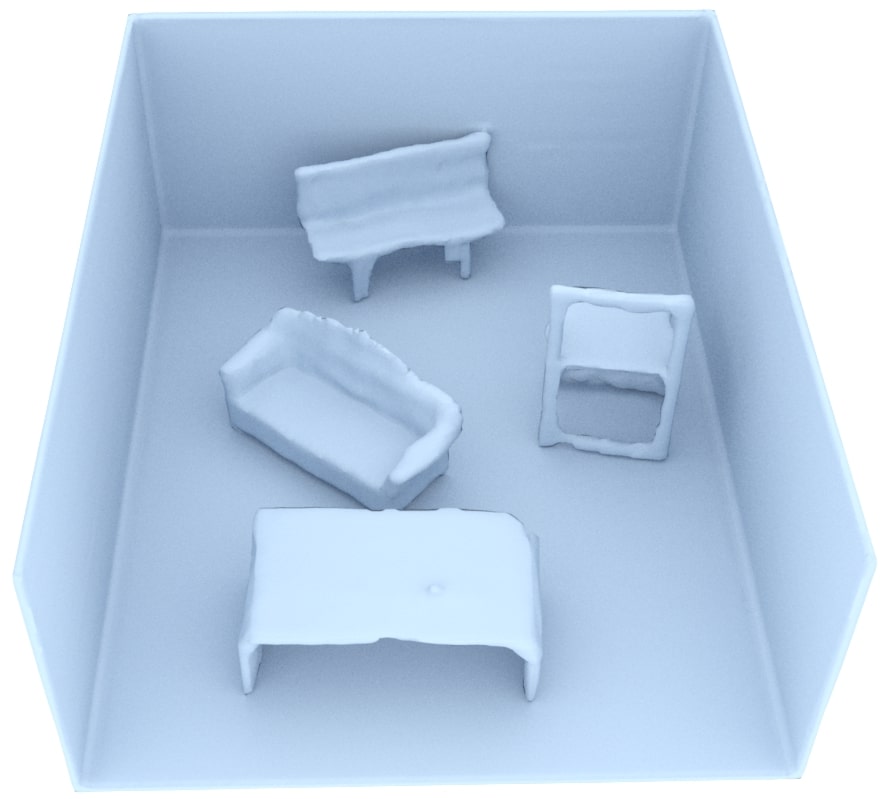}} &   \centered{\includegraphics[width=0.17\textwidth]{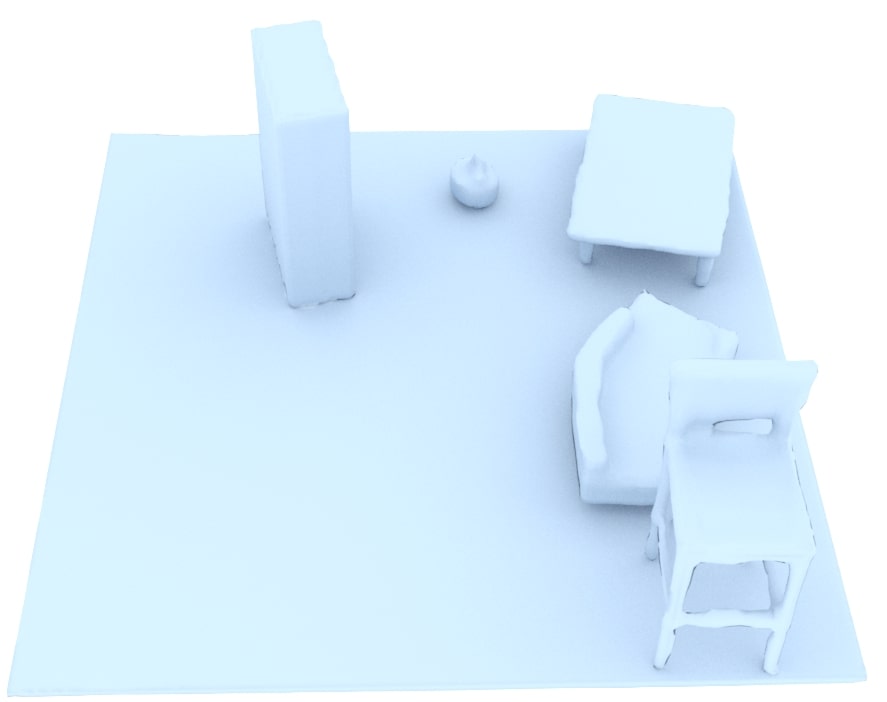}} &  
  \centered{\includegraphics[width=0.23\textwidth]{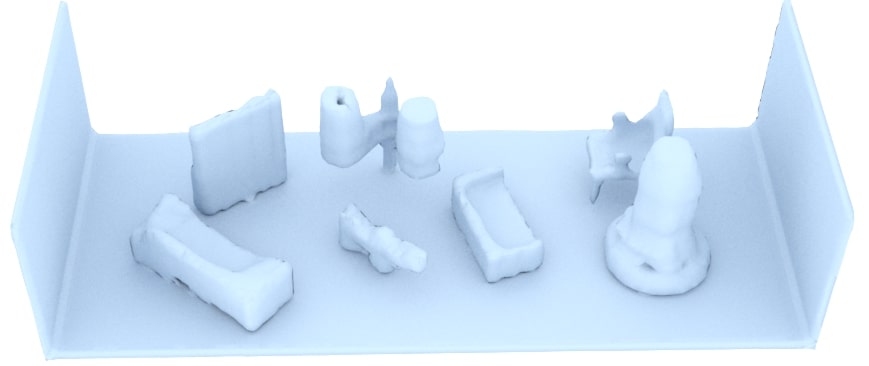}} &  \centered{\includegraphics[width=0.11\textwidth]{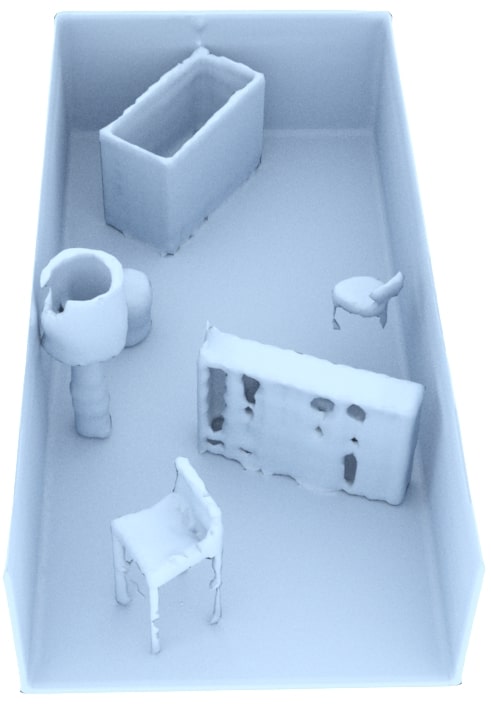}} \\
  \centered{\rotatebox[origin=c]{90}{Ours}} & 
  \rotatebox[origin=c]{90}{(5D with PE)} &
  \centered{\includegraphics[width=0.15\textwidth]{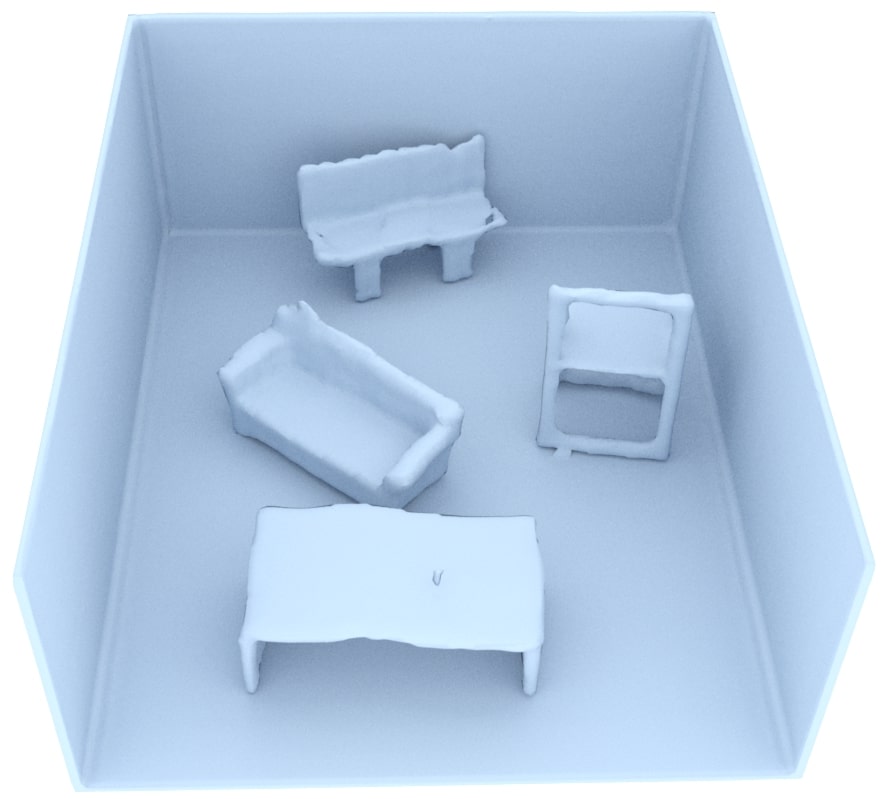}} &  \centered{\includegraphics[width=0.18\textwidth]{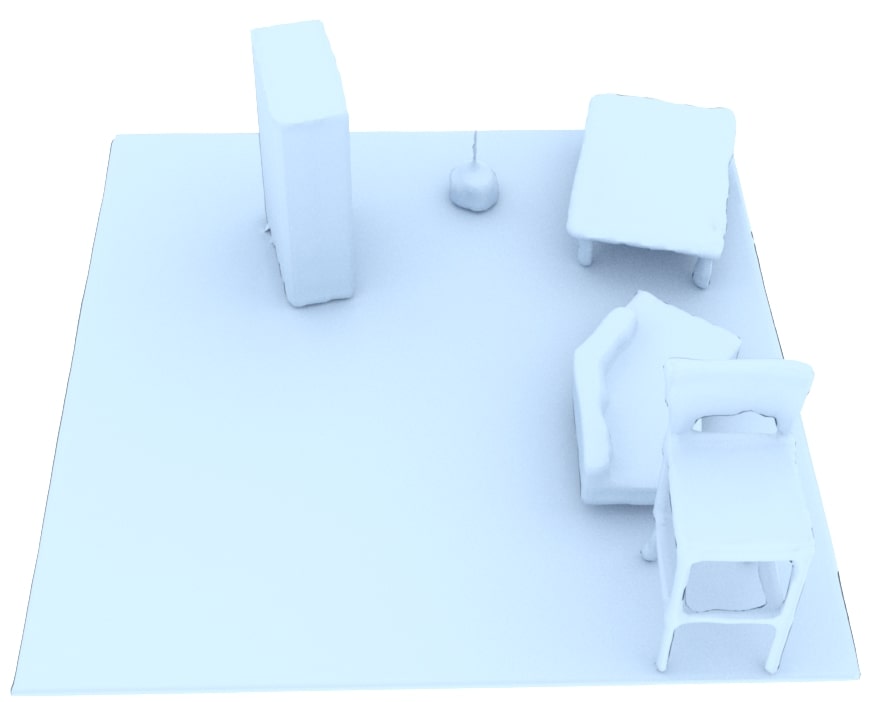}} &  
  \centered{\includegraphics[width=0.23\textwidth]{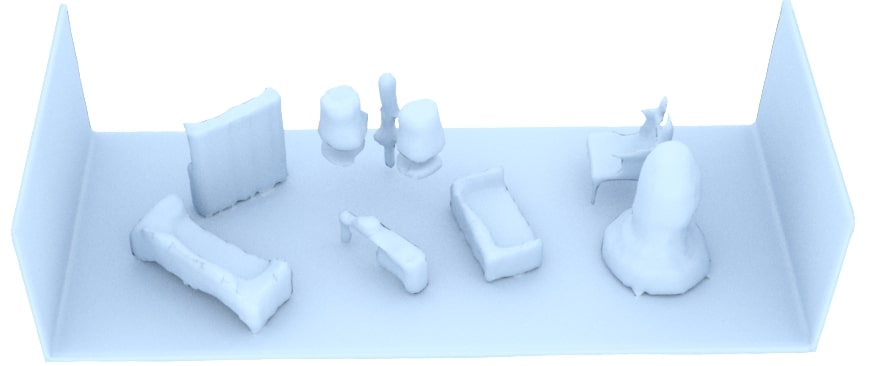}} &  \centered{\includegraphics[width=0.115\textwidth]{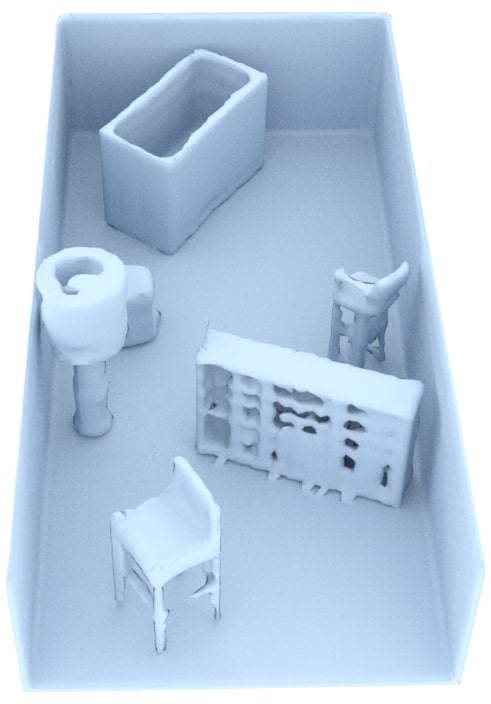}} \\
\end{tabular}
\end{center}
\caption{\textbf{Scene-level reconstruction on synthetic rooms.} Qualitative comparison of synthetic indoor scene reconstruction from point clouds.}
\label{fig:scene}
\end{figure*}

\subsection{Scene-Level Reconstruction}

\begin{table}[b!]
\centering
\resizebox{0.475\textwidth}{!}{%
\begin{tabular}{|l|cccc|}
\hline
  & IoU & Chamfer- & Normal & F-score \\
 & & $L_1$ & C. & \\
\hline\hline
\textbf{Without PE} & & & & \\
ONet \cite{mescheder2018occupancy}& 0.475 & 0.203 & 0.783 & 0.541 \\
ConvONet (3C) \cite{peng2020convolutional} & 0.789 & 0.044 & 0.902 & 0.950 \\
ConvONet (3C +  & 0.816 & 0.044 & 0.905 & 0.952 \\
 $32^3$ grids) \cite{peng2020convolutional} &  &  &  & \\
Ours (3D) & 0.795 & 0.043 & 0.907 & 0.954\\
Ours (5D) & 0.791 & 0.043 & 0.905 & 0.955 \\
Ours (7D) & 0.810 & \textbf{0.042} & 0.909 & 0.957\\
Ours (3C + 2D) & \textbf{0.837} & \textbf{0.042} & 0.910 & 0.958 \\
Ours (3C + 4D) & 0.831 & 0.044 & 0.906 & 0.953\\
\hline
\textbf{With PE} & & & & \\
ConvONet (3C) \cite{peng2020convolutional} & 0.797 & 0.046 & 0.902 & 0.946 \\
Ours (3D)  & 0.814 & \textbf{0.042} & 0.910 & 0.958 \\
Ours (5D) & 0.800 & \textbf{0.042} & \textbf{0.912} & \textbf{0.960} \\
Ours (7D) & 0.819 & 0.043 & 0.910 & 0.957 \\
Ours (3C + 2D) & 0.797 & 0.043 & 0.908 & 0.959 \\
Ours (3C + 4D) & 0.831 & 0.043 & 0.910 & 0.956 \\
\hline
\end{tabular}}\\
\footnotesize{PE = positional encoding. C = canonical planes. D = dynamic planes. SL = similarity loss.}
\caption{\textbf{Scene-level reconstruction on synthetic rooms.} Our results on the synthetic indoor scene dataset.}

\label{tab:scene_result}
\end{table}

For the scene-level experiment, we uniformly sample 10,000 points from the ground truth meshes as input and apply Gaussian noise with a standard deviation of 0.05. During training, we query the occupancy probability of 2048 points. 
We set the plane resolution $128^2$ and use U-Net with a depth of 5.
The batch size during training is set to 32 for all experiments with 3 planes, while
a batch size of 16 is used for experiments with 5 and 7 planes to accommodate the higher GPU memory requirement. The models are trained for at least 500,000 iterations. 

We train our models for synthetic indoor scene dataset by applying the similarity loss (Eq. \ref{eq:simloss}) and disabling after 20,000 iterations, which enables more robust training in our experiments.
The reason for doing so is: we find several of our runs without the similarity loss initialization have considerably higher training loss and lower validation score. We observe those models do not predict one of the canonical planes and have planes angled less than $45^\circ$. Our speculation of this occurrence is because the scene dataset has similar global structures of rectangular shapes, it is difficult for our plane predictor networks to recover from bad minimas when the plane prediction is not governed by the similarity loss.

As shown in Table~\ref{tab:scene_result}, our models achieve better accuracy in all metrics. Moreover, it can be seen from Fig.~\ref{fig:scene} that our models preserve better fine-grained details than the baseline methods.

%
\section{Conclusion}
In this work, we introduced Dynamic Plane Convolutional Occupancy Networks, a novel implicit representation method for 3D reconstruction from point clouds. We proposed to learn dynamic planes to form informative features. We observe that 3 canonical planes are always predicted, and the symmetric property of objects are implicitly encoded. We also find that enforcing a similarity loss on the predicted plane normals considerably improves the performance on unseen object poses. In future work, we plan to assess the theoretical support for the dynamic plane prediction. \\ \\

{\small
\bibliographystyle{ieee_fullname}
\bibliography{egbib}
}

\end{document}